\documentclass[manuscript,screen,acmsmall,nonacm]{acmart}

\AtBeginDocument{%
  }

\usepackage{multirow}
\usepackage[table,xcdraw]{xcolor}

\definecolor{LightGray}{gray}{0.94}

\definecolor{humanAIcollaboration}{HTML}{4e79a7}
\definecolor{AIonly}{HTML}{e15759}
\definecolor{humanOnly}{HTML}{59a14f}
\definecolor{IDontKnow}{HTML}{b07aa1}

\definecolor{NotAtAllAcceptable}{HTML}{F8766D}
\definecolor{RatherUnacceptable}{HTML}{B79F00}
\definecolor{NeutralAssessment}{HTML}{00BA38}
\definecolor{RatherAcceptable}{HTML}{00BFC4}
\definecolor{DefinitelyAcceptable}{HTML}{619CFF}
\definecolor{IDontKnow}{HTML}{F564E3}

\usepackage{subcaption}
\usepackage{tcolorbox}

\theoremstyle{plain}

\theoremstyle{definition}

\newcommand{\ourtitle}{Reduced AI Acceptance After the Generative AI Boom: Evidence From a Two-Wave Survey Study}

\usepackage{xcolor}

\definecolor{DarkGreen}{rgb}{0.075,0.375,0.075}
\definecolor{DarkRed}{rgb}{0.5,0.1,0.1}
\definecolor{DarkBlue}{rgb}{0.1,0.1,0.5}
\definecolor{Gray}{rgb}{0.2,0.2,0.2}

\newcommand{\DESATUZH}{DESAT process of the University of Zurich (for details see \url{https://www.dsi.uzh.ch/en/research/projects/archive/strategic/desat.html})}

\definecolor{wave1}{HTML}{E6F3FF}  
\definecolor{wave2}{HTML}{FFF0E6}  

\usepackage{tikz}
\usepackage{subcaption}
\usepackage{enumitem}

\usepackage{booktabs}
\usepackage{makecell}

\usepackage[flushleft]{threeparttable}
\usepackage{transparent}

\begin{document}

\title{\ourtitle}

\author{Joachim Baumann}
\authornote{Corresponding author}
\email{joachim.baumann@unibocconi.it}
\orcid{0000-0003-2019-4829}
\affiliation{%
  \institution{University of Zurich and Bocconi University}
}

\author{Aleksandra Urman}
\email{urman@ifi.uzh.ch}
\orcid{0000-0003-3332-9294}
\affiliation{%
  \institution{University of Zurich}
}

\author{Ulrich Leicht-Deobald}
\email{ulrich.leicht-deobald@tcd.ie}
\orcid{0000-0003-4554-7192}
\affiliation{%
  \institution{University of Dublin}
}

\author{Zachary J. Roman}
\email{zacharyjoseph.roman@uzh.ch}
\orcid{0000-0003-0246-4952}
\affiliation{%
  \institution{University of Zurich}
}

\author{Anik\'{o} Hann\'{a}k}
\email{hannak@ifi.uzh.ch}
\orcid{0000-0002-0612-6320}
\affiliation{%
  \institution{University of Zurich}
}

\author{Markus Christen}
\email{markus.christen@dsi.uzh.ch}
\orcid{0000-0001-8378-9366}
\affiliation{%
  \institution{University of Zurich}
}

\renewcommand{\shortauthors}{Baumann et al.}

\begin{abstract}
The rapid adoption of generative artificial intelligence (GenAI) technologies has led many organizations to integrate AI into their products and services, often without considering user preferences. Yet, public attitudes toward AI use, especially in impactful decision-making scenarios, are underexplored. Using a large-scale two-wave survey study ($n_{\text{wave 1}}=1514$, $n_{\text{wave 2}}=1488$) representative of the Swiss population, we examine shifts in public attitudes toward AI before and after the launch of ChatGPT.
We find that the GenAI boom is significantly associated with reduced public acceptance of AI (see Figure~\ref{fig:average_acceptance_comparison_by_scenario}) and increased demand for human oversight in various decision-making contexts.
The proportion of respondents finding AI ``not acceptable at all'' increased from 23\% to 30\%, while support for human-only decision-making rose from 18\% to 26\%.
These shifts have amplified existing social inequalities in terms of widened educational, linguistic, and gender gaps post-boom.
Our findings challenge industry assumptions about public readiness for AI deployment and highlight the critical importance of aligning technological development with evolving public preferences.
\end{abstract}


\begin{teaserfigure}
\centering
\includegraphics[width=0.77\textwidth]{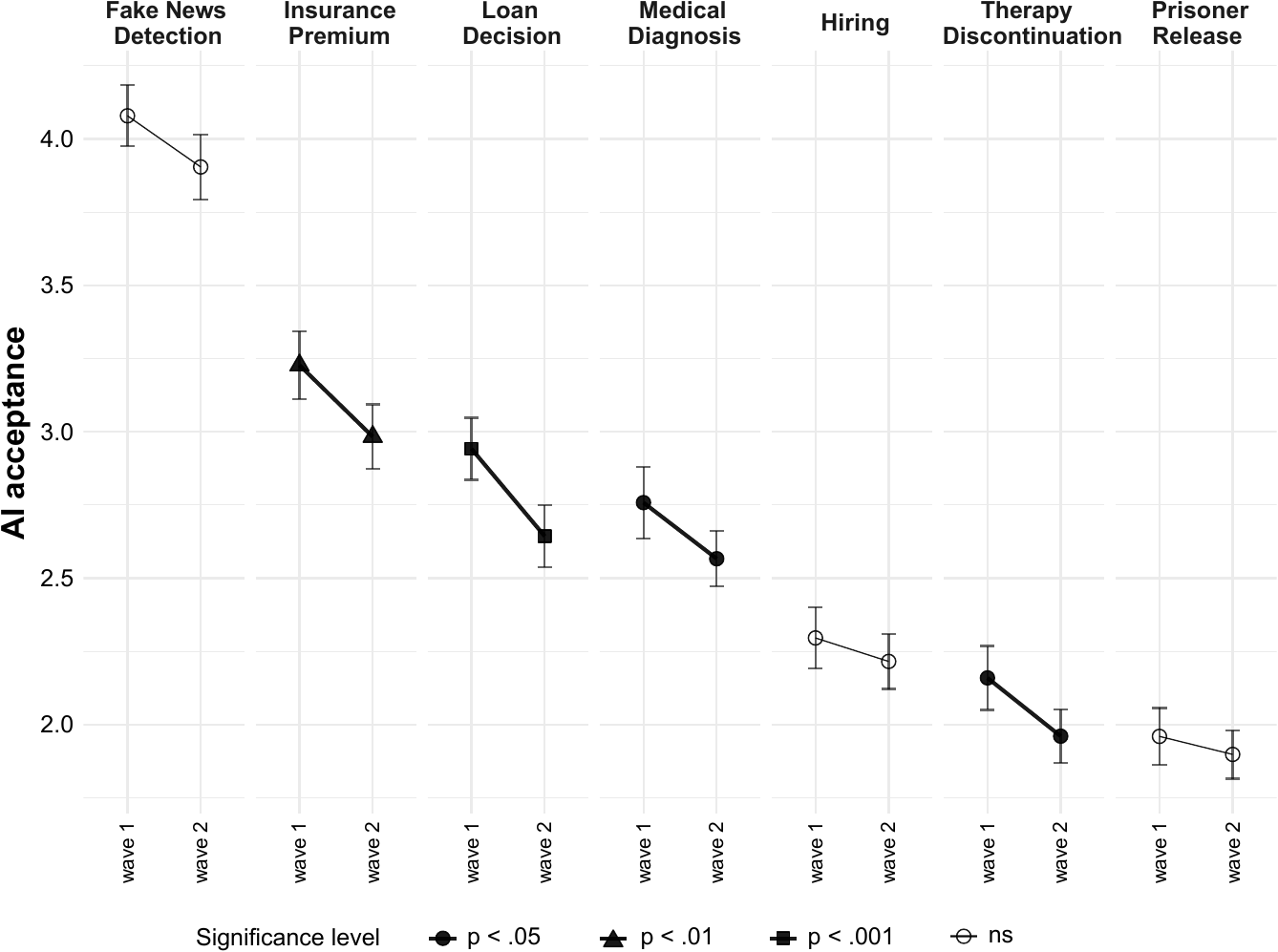}
\caption{
The generative AI boom that occurred between two survey waves (January-February 2022 and July-August 2023) is associated with a significant decrease in AI acceptance for four out of seven decision-making scenarios among a representative Swiss sample.
The error bars represent 95\% confidence intervals (CI) and significance levels indicate statistically significant differences across survey waves as determined by a two-sample design-based t-test.
}
\label{fig:average_acceptance_comparison_by_scenario}
\end{teaserfigure}

\maketitle

\section{Introduction}

Artificial intelligence (AI) has become ubiquitous in our daily lives, appearing even in unexpected places such as pillows and toothbrushes~\citep{clayton2024ces}.
The recent proliferation of generative AI (GenAI) technologies has driven adoption across various domains, from customer service and search engines to health consultations~\citep{Hall2024,Reid2024,BingGenSearch2024,lautrup2023heart}.
This rush to incorporate GenAI into products and services is often motivated by a fear of falling behind, leading organizations to hastily adopt AI technologies without fully considering public acceptance.
Industry professionals believe in the positive impact of AI~\citep{reuters2024futureprofessionals} and often also believe (e.g., in media~\citep{Blassnig2024}) that customers expect the implementation of new, AI-powered tools.
However, this perception may diverge from people's preferences, potentially harming organizations' legitimacy due to the widening gap between imagined and real users~\citep{Strikovic2024}.

Furthermore, with the widespread adoption of AI and its growing capabilities, governments around the world have started issuing regulations to manage its use. A critical element of these regulatory efforts is the clear specification of the required levels of human oversight and control over AI systems.
The European Union's Artificial Intelligence Act (EU AI Act), which came into force on August 1, 2024, mandates specific human oversight measures depending on the risk that AI systems may cause~\citep{EU2024AIAct}.
Regulations have also been proposed in other countries, such as the U.S. executive order on AI, published on October 30, 2023~\citep{biden2023executive}.

Therefore, although in today's AI race organizations may try to find more profitable AI use cases~\citep{reuters2024futureprofessionals}, they may overlook the opinions and acceptance levels of the general population regarding their use of AI.
Moreover, the broader implications of this AI adoption, such as exacerbating existing inequalities -- including pronounced distrust of health systems among women and general technological inequality -- remain largely underexplored~\citep{acemoglu2002technology,Garcia2018genderinequality,Bhandari2019,edelmantrust2019}.
Moreover, despite ongoing regulation efforts, we still lack a concrete understanding of public opinion on the appropriate levels of human control required for AI systems in different organizational application contexts.
This gap underscores the need for research that quantifies laypeople's perceptions and provides actionable insights for policymakers.~\looseness=-1

\subsection{Contributions.}

In this study, we aim to investigate the public's acceptance of AI use in different decision-making scenarios and their corresponding preferences for human control through a representative two-wave survey study in Switzerland.
Our paper makes three important theoretical contributions.

First, our research substantiates that public opinions shift over time, particularly during a technological boom.
Pre-GenAI boom studies have quickly become outdated, as public discourse has shifted tangibly~\citep{modhvadia2023people}.
Our survey design uniquely addresses this by running the same questionnaire in two waves with a temporal gap that captures the shift in public sentiment towards AI in times of technological disruption.
Additionally, analyses of detailed individual characteristics reveal increased demographic disparities along the lines of gender, education, and language regions over time.

Second, our paper contributes by examining AI acceptance in Switzerland, a country known for cultural diversity (being at the intersection of German, French, and Italian cultural European spaces) and high levels of innovation and digital literacy~\citep{MARXT2013CHInnovation}.
This diversity makes Switzerland an ideal setting to explore how contextual differences may shape public opinion on AI. Previous research has mainly focused on the U.S. context~\citep{kaufmann_task-specific_2023}, and prior work suggests one cannot simply generalize results from one country to others since AI acceptance varies widely across cultural and national contexts~\citep{kaufmann_task-specific_2023,fletcher_what_2024}.
Furthermore, existing studies predominantly relied on non-representative samples, such as university students, Amazon Mechanical Turk workers, or medical clinic visitors, which may not accurately reflect the general population~\citep{kelly_what_2023,chen2022acceptance,yilmaz2023student,HO2023healthcare}.
To overcome these limitations, we employed a two-wave survey design representative with respect to age, gender, language, education, and political orientation to capture more reliable insights into public sentiment in Switzerland toward different AI use cases.

Finally, we contribute to the literature on human control in AI by offering a fine-grained conceptualization of AI versus human control.
Prior research has found that human involvement remains essential and that people are often hesitant to give machines
full control~\citep{Tolmeijer2022Capable,Sundar2020}.
However, European regulation only provides a high-level description of human oversight requirements~\citep{EU2024AIAct}, which does not reflect the complexity of organizational realities in monitoring AI~\citep{Wang2021,Bansal2021Does,Zhang2022YouComplete}.
Building on prior work on algorithm aversion and human-AI collaboration~\citep{Green2019,Dietvorst2018,Cheng2023}, we address this complexity by examining shifts in public preferences for different levels of human control in AI systems over time.
In particular, we differentiate between four levels of human control: (a) human-only decisions, (b) human decisions with AI assistance, (c) AI decisions with possible human intervention, and (d) fully autonomous AI decisions.
This can help regulators establish clearer recommendations on the appropriate levels of human control across different use cases.

\section{Related work on contextual AI acceptance, human control, and the GenAI boom}

\subsection{Contextual AI acceptance}

In this paper, we focus on \textit{contextual} AI acceptance, or the extent to which an individual finds it acceptable to use AI \textit{in a given domain}.
Recent reviews of the literature on AI acceptance~\citep{kelly_what_2023,kaufmann_task-specific_2023} have shown that most studies to date have focused either on the acceptance of AI in general or on AI in single use cases and/or industries, such as education~\citep{kim_my_2020}, healthcare~\citep{antes_exploring_2021}, or public services~\citep{schmager_what_2023}.
While such studies provide important insights about given usage scenarios, the few studies that examined AI acceptance across usage contexts from a comparative perspective show that AI acceptance varies widely across decision-making scenarios and industries~\citep{castelo_task-dependent_2019,kaufmann_task-specific_2023,modhvadia2023people,jussupow_why_2020,araujo_ai_2020}.
For example, medicine has been consistently found to be among the domains in which the acceptance of AI tends to be the lowest~\citep{kaufmann_task-specific_2023,jussupow_why_2020}.
This might be related to the fact that medical decisions are highly consequential, and more consequential decision-making scenarios have been linked to lower contextual AI acceptance~\citep{castelo_task-dependent_2019}.

It is important to note that acceptance and trust are distinct concepts in the technology adoption literature.
While trust can be an antecedent or predictor of acceptance~\citep{kelly_what_2023}, acceptance itself encompasses the behavioral intention or willingness to use, buy, or try a technology~\citep{Davis_1989}.
In our study, we build on acceptance models from the literature~\citep{Davis_1989,Gursoy_et_al_2019}, adapted to our use case, i.e., to investigate users' willingness to accept or object to AI usage in various decision-making scenarios.

While AI acceptance varies across contexts, comparative -- that is, cross-industry/cross-usage scenario -- evaluations of AI acceptance are rare~\citep{kelly_what_2023}. This limits our ability to draw generalizable conclusions about AI acceptance across different scenarios.
While in principle, one could make such conclusions by, for example, comparing acceptance levels observed in the studies focused on specific single scenarios, in practice, such direct comparisons are often not possible for two main reasons.
First, individual studies are conducted at different points in time, and AI acceptance can change from one point to another.
Second, the studies rely on different samples, often drawing conclusions not from representative samples of the population but from specific social groups such as students, business professionals, or even Amazon Mechanical Turk workers~\citep{kelly_what_2023,kaufmann_task-specific_2023}.
Therefore, we focus on \textit{contextual} acceptance of AI rather than more general AI acceptance.

\subsection{Human control of AI}

The EU AI Act underscores the necessity of \textit{human oversight} for \textit{high-risk} AI systems, stating that:
``High-risk AI systems shall be designed and developed in such a way, including with appropriate human-machine interface tools, that they can be effectively overseen by natural persons during the period in which they are in use.''~\citep[Article 14]{EU2024AIAct}.
This emphasis on human oversight is particularly crucial in high-impact decision-making contexts across various domains, such as healthcare~\citep{Aoki2021Theimportanceof,Zanzotto2019HilAI,Gronsund2020Augmentingthealgorithm,Mosqueira2023Humanintheloop,Wu2022survey,Benedikt2020Human}.
In clinical decision support systems, for example, AI can provide valuable assistance in diagnosing patients or flagging potential issues, but human involvement remains essential to ensure that the AI's recommendations are correctly interpreted and applied~\cite{Wang2021Brilliant}.

Human control in AI systems is often conceptualized along a spectrum, ranging from full human control to full AI autonomy~\cite{Kim2023OneAI}. At one extreme, humans make all decisions without AI input; at the other, AI systems operate entirely autonomously without human intervention. Between these two poles, various collaborative forms exist, such as AI-assisted human decisions or AI-generated decisions subject to human oversight~\cite{abbass2019social,Takayama2015Telepresence,rammert2008action}. These models are also commonly referred to as \textit{human-in-the-loop} -- where the human retains primary decision-making authority -- and \textit{human-on-the-loop} -- where the human supervises and can intervene to correct AI-driven decisions~\cite{fischer2021loop}.~\looseness=-1

The degree of human control over AI significantly shapes how users perceive the role of AI in decision-making, with studies showing a general preference for higher levels of human involvement~\cite{Sundar2020,Babamiri2022Insights,mays2022ai}. Understanding the nuances of this preference, particularly in high-impact scenarios, is crucial for designing AI systems that are trusted and widely accepted.
Prior experimental studies have examined various dimensions of human-algorithm interaction, including outcome modification~\citep{Dietvorst2018}, process versus outcome control~\citep{Cheng2023}, and the principles that should guide algorithm-in-the-loop decision making~\citep{Green2019}. These works highlight the importance of accounting for different levels of human control when designing AI systems. Our study complements this literature by examining how public perceptions have shifted regarding different human-AI collaboration models in decision-making contexts.

\subsection{Generative AI boom}

Attitudes toward technologies are not static and may change over time, oftentimes nurtured by disruptive changes~\citep{Karahanna1999}.
In November 2022, the launch of ChatGPT attracted worldwide attention and marked a crucial moment in the public discourse surrounding AI -- with then unknown effects on the public's acceptance towards AI~\citep{modhvadia2023people}.
The widespread accessibility of ChatGPT and the capabilities of its underlying large language model (LLM)\footnote{See OpenAI's technical report on their latest LLM~\citep{openai2024gpt4technicalreport}.}, along with some other highly capable generative AI models, such as Stable Diffusion~\citep{rombach2022stablediffusion} or DALL-E 2~\citep{ramesh2022dalle2}, that were released at a similar time, sparked intense discussions on the potential benefits and challenges of AI~\citep{Wu2023ChatGPT,lund2023chatting,Kasneci2023ChatGPT}.
This surge in interest is reflected across various domains, from everyday conversations to extensive media coverage and governmental deliberations, and has been referred to as the \textit{generative AI boom}~\citep{wikiAIboom,Knight2023AIboom}.~\looseness=-1

\begin{figure*}[h!]
\centering
\includegraphics[width=1\textwidth]{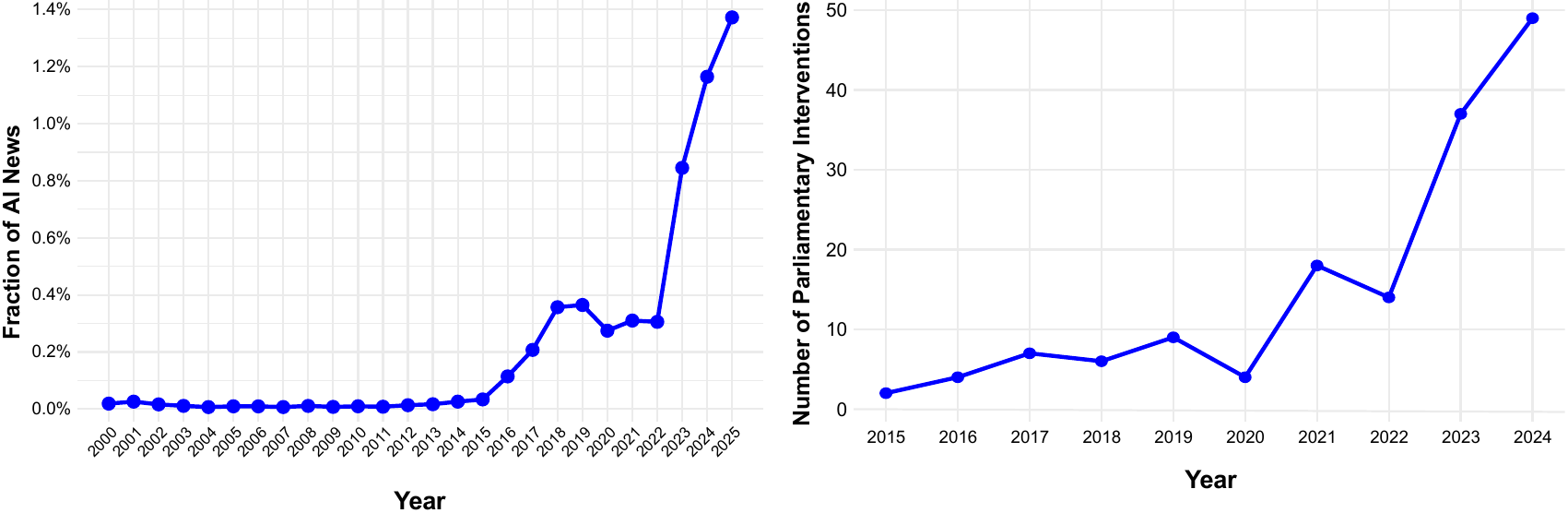}
\caption{The generative AI boom in Swiss media and politics.
The left panel shows the fraction of AI news in all Swiss news from 2000 to 2025 that mention AI (weighted by the total number of news per year).
The left panel shows the number of parliamentary interventions on AI from 2015 to 2024 involving AI-related terms.
}
\label{fig:aiboom}
\end{figure*}

As shown in Figure~\ref{fig:aiboom} (left), the generative AI boom has led to a drastic increase in the proportion of news articles about AI in Switzerland.
The data\footnote{This data was retrieved from Factiva (c.f. \url{https://global.factiva.com}), an international news database produced by Dow Jones that contains 187 different Swiss news outlets. We queried Swiss media outlets using the search terms ``Künstliche Intelligenz,'' ``Intelligence artificielle,' and ``Intelligenza artificiale,'' representing the translation of AI into German, French, and Italian -- the three official languages in Switzerland in which most news media are published.} indicates that coverage has tripled over the period, highlighting growing media attention towards AI.

Similarly, the AI boom has also led to a parallel growth in political activity related to AI advancements and implications in Switzerland, as shown in Figure~\ref{fig:aiboom} (right).
This data\footnote{This data was retrieved from Curia Vista, a database containing all parliamentary proceedings of the Federal Assembly of Switzerland since 1995 -- c.f. \url{https://www.parlament.ch/en/ratsbetrieb/curia-vista}.} reflects a major rise in the number of parliamentary interventions involving AI-related terms, underscoring the political response to the evolving AI landscape.

\noindent
Public attitudes towards the use of AI, as well as the need for human oversight, are likely influenced by shifts in public discourse following major technological advancements~\citep{zhang2019artificial,funk2020publics,bdcc7010035}.
Notably, the launch of ChatGPT has been a significant milestone in AI development. With this context in mind, we aim to explore the following research questions:
\begin{tcolorbox}[colframe=gray!90, colback=gray!20, fonttitle=\bfseries, boxrule=0.5pt, left=3pt, right=3pt, top=1pt, bottom=3pt, boxsep=5pt]
\textit{
\textbf{
\begin{enumerate}[left=0pt, topsep=2pt, itemsep=5pt, parsep=0pt, label=RQ1\alph*]
    \item How did the AI acceptance change with the generative AI boom?
    \item How did the preferences for human control of AI change with the generative AI boom?~\looseness=-1
\end{enumerate}
}
}

\end{tcolorbox}
\noindent
Since AI acceptance and human control are highly contextual~\citep{castelo_task-dependent_2019,kaufmann_task-specific_2023,modhvadia2023people,jussupow_why_2020,araujo_ai_2020}, we investigate both research questions across several specific usage scenarios.

\subsubsection{AI acceptance and individual characteristics}
AI usage scenarios and their characteristics are only one factor explaining the differences in AI acceptance rates. However, even within the same usage scenario, AI acceptance can also vary based on people's individual characteristics, including socio-demographic attributes~\citep{kelly_what_2023,mendez-suarez_factors_nodate}.

\textit{Gender} consistently emerges as a significant predictor of AI acceptance, with studies showing that women tend to be more skeptical of AI, perceive it as less useful, and see fewer personal and societal benefits in its adoption~\citep{mendez-suarez_factors_nodate,araujo_ai_2020,sartori_minding_2023,cho_dual_2024,horowitz_what_2021,Gonzalez2024Sociodemographic}. These findings align with broader research on gender differences in technology acceptance generally, where similar patterns have been observed for digital technologies~\citep{cai_gender_2017,venkatesh_why_2000}. Hence, we anticipate that gender will be significantly associated with the respondents' AI acceptance in our study as well, with women exhibiting lower acceptance towards AI use generally. At the same time, we expect the significance and/or the effect size to be different across AI use scenarios, in line with the findings of ~\citet{araujo_ai_2020} who showed that gender differences in the perceptions of AI-based decision-making systems' risk,usefulness, and fairness vary between different decision-making domains.

\textit{Age} is another demographic factor linked to lower acceptance of AI and technology in general, with older individuals generally holding more negative attitudes towards AI~\cite{araujo_ai_2020, mendez-suarez_factors_nodate,Gonzalez2024Sociodemographic}.
Additionally, higher levels of \textit{education} and \textit{familiarity with AI} technologies have been shown to correlate positively with AI acceptance and trust in AI decision-making~\citep{araujo_ai_2020}.~\looseness=-1

Given these established group-level differences, we aim to explore potential changes in these dynamics during the rise of generative AI:
\begin{tcolorbox}[colframe=gray!90, colback=gray!20, fonttitle=\bfseries, boxrule=0.5pt, left=3pt, right=3pt, top=1pt, bottom=3pt, boxsep=5pt]
\textit{
\textbf{
\begin{enumerate}[left=0pt, topsep=2pt, itemsep=3pt, parsep=0pt, label=RQ2:]
    \item How have group differences in AI acceptance, based on gender, age, education, language regions, digital skills, and AI knowledge, evolved with the generative AI boom?~\looseness=-1
\end{enumerate}
}
}
\end{tcolorbox}

\section{Method}

\subsection{Data collection}
\label{ssec:survey_execution}

We conducted two surveys, the first in January and February 2022, before the GenAI boom, and the second 17 months later in July and August 2023, after the GenAI boom.
Both surveys were distributed by a professional survey company\footnote{gfs Bern; c.f. \url{https://www.gfsbern.ch/en/}} considering quotas for age, gender, and language region to ensure the representativeness of the adult Swiss population based on data from Switzerland's Federal Statistical Office~\citep{bfs2023,bfs2020}.

The final samples consist of 1514 1st wave respondents and 1488 2nd wave respondents (more details in Table~\ref{tab:survey_overview}).
In total, the 1st survey wave questionnaire was accessed 2318 times. 566 persons (24\%) dropped out of the survey; another 238 persons (10\%) were excluded due to failure in the attention check.
Of the 2056 persons that started the survey wave 2, about 22\% dropped out, and about 8\% were excluded after failing the attention check.\footnote{
The attention check consisted of a question about favorite colors with embedded instructions to select both ``Red'' and ``Green'' options, regardless of actual preference. Participants who failed to follow these specific directions were excluded from the survey.
}

\begin{table*}[thb]
\centering
\caption{Survey execution and sample sizes.}
\begin{tabular}{lcc}
\toprule
 & \textbf{Wave 1} & \textbf{Wave 2} \\
 & (pre GenAI boom) & (post GenAI boom) \\
\midrule
Survey period & Jan 20 -- Feb 12, 2022 & Jul 26 -- Aug 22, 2023 \\
Total participants\textsuperscript{a} & 2318 & 2056 \\
Participants dropped out & 566 & 443 \\
Excluded answers\textsuperscript{b} & 238 & 125 \\
\textbf{Final sample size} & \textbf{1514} & \textbf{1488} \\
\bottomrule
\end{tabular}
\label{tab:survey_overview}
\vspace{0.1cm}
\begin{tabular}{p{0.75\linewidth}}
\small
\textsuperscript{a} This includes all participants who started filling out the survey but excludes those who were not admitted due to unfulfilled demographic quotas. \\
\small
\textsuperscript{b} Both surveys included an attention question, and we only considered data from participants who answered correctly. \\
\end{tabular}
\end{table*}

We inspected ethical and privacy issues based on the
\DESATUZH.
Since we collected the data anonymously and none of the questions involved specifically sensitive issues, the study was evaluated as ``low risk'' and did not require approval by an ethics board.
This is also in line with the checklist of the ethical committee of the lead author's institution.

\subsection{Questionnaire}

\subsubsection{Demographics}
The questionnaire structure was designed by the authors of this manuscript.
Participants could choose to fill out the questionnaire in any of the official languages at the national level, i.e., German, French, or Italian.
Furthermore, the beginning of the questionnaire included a set of socio-demographic questions (gender, age, language, education, and political orientation).~\looseness=-1

\paragraph{Digital skills and AI literacy.}
Next, we evaluated the respondents' digital skills.
Measuring digital skills and AI literacy presents challenges due to the rapidly evolving technological landscape.
For this study, we utilized three different scales in the first wave of our survey to assess participants' competencies in these areas.
The first scale from~\citet{Hargittai2012SurveyMeasures}, evaluates respondents' familiarity with advanced digital concepts, including advanced search, PDF, spyware, wiki, cache, and phishing (wave 1: Cronbach's $\alpha$=0.88, $\mu=3.29$, $\sigma=0.94$; wave 2: Cronbach's $\alpha$=0.87, $\mu=3.34$, $\sigma=0.91$).
The second scale, which was developed by~\citet{Franke2019ATI}, measures affinity for technology interaction through a 9-item questionnaire, reflecting individuals' propensity to engage with technology (wave 1: $\mu=3.29$, $\sigma=0.94$).
Additionally, we introduced a scale for AI literacy, asking participants to rate their understanding of the terms ``artificial intelligence,'' ``machine learning,'' ``neural networks,'' and ``deep learning'' (wave 1: Cronbach's $\alpha$=0.89, $\mu=3.65$, $\sigma=0.98$).~\looseness=-1

The strong correlations observed among these scales (Spearman rank correlations ranging from 0.45 to 0.56, all with $p < 0.001$)
support their reliability and alignment.
Therefore, we used only the literacy scale by \citet{Hargittai2012SurveyMeasures}, which is well-established in the literature, for the second wave and based all analyses in this manuscript on this measure to ensure consistency and reliability in measuring digital skills.

In the second survey wave, we additionally asked respondents about their familiarity with ChatGPT (see answer options in Table~\ref{tab:ChatGPT_awareness}).

\subsubsection{AI primer}

The concept of AI lacks of a universally accepted definition~\citep{Wang2019DefiningAI,modhvadia2023people,pataranutaporn2023influencing}.
Thus, it is crucial to appropriately prime study participants before they respond to the questionnaire.
By ensuring that participants share a common understanding of AI, we can mitigate variability in interpretation and obtain more consistent responses.
Therefore, we primed the participants with the following explanation, ensuring a common definition of AI and its use in decision-making.~\looseness=-1
\begin{tcolorbox}[colframe=gray!90, colback=gray!20, title=AI primer, fonttitle=\bfseries, boxrule=0.5pt, left=5pt, right=5pt, top=0pt, bottom=0pt, boxsep=3pt]
When we talk about ``artificial intelligence'' below, we mean the following:
The term ``artificial intelligence'' (AI) refers to computer software that learns from data and is then able to make decisions or predictions.
Examples include:
\begin{itemize}[left=0pt, topsep=0pt, itemsep=0pt, parsep=0pt]
    \item Digital assistants such as ``Alexa'' from Amazon or ``Siri'' from Apple
    \item Systems that enable self-driving cars
    \item Systems that make personalized recommendations (e.g. Amazon)
    \item Systems that control robots
    \item Systems that moderate content on social media platforms (e.g. detection of hate speech)
    \item Systems that enable facial recognition (e.g. for access control)
\end{itemize}
In this survey, ``artificial intelligence'' (AI) refers to those AI systems that are in use today and their obvious further developments. It does not refer to unrealistic systems that surpass human intelligence, such as ``Skynet'' in the Terminator movies, robots like in the movie ``I Robot'' or ``HAL 9000'' in the movie ``2001 A Space Odyssey''.
Please base the following questions on this understanding of AI. 
\end{tcolorbox}
The primer was specifically designed to orient participants toward existing contemporary AI applications rather than speculative or futuristic conceptions. While alternative definitions of AI exist (see Section~\ref{ssec:Limitations_and_future_work} for a detailed discussion), for the remainder of this section, we proceed on the assumption that questionnaire respondents were successfully primed and thus share a common understanding of AI.

\subsubsection{AI acceptance and human control}
In the main part of the survey, we assessed the AI acceptance and human control conditions of using AI for automating decision-making, which is motivated by prior work~\citep{araujo_ai_2020}.
To capture contextual variations, we consider seven paradigmatic applications, which are listed in Table~\ref{tab:scenarios}.
All scenarios concern already existing or likely real-world AI applications (i.e., fake news detection, insurance premium pricing, loan decisions, medical diagnosis, hiring decisions, therapy discontinuation, and prisoner release decisions).

For AI acceptance, we asked the question \textit{``How acceptable would it be for you if this decision was made by an AI system?''} with standard five-point Likert scales plus an additional option for respondents to indicate that they don't know.
Hence, respondents could choose one of the following 6 answer possibilities:
(1) \textit{Not acceptable at all},
(2) \textit{rather not acceptable},
(3) \textit{neutral assessment},
(4) \textit{rather acceptable},
(5) \textit{definitely acceptable},
(6) \textit{I don't know}.

For human control, we asked survey participants to ``\textit{Please assess what would be the optimal way to make the following decisions,}'' by selecting one of the options listed in Table~\ref{tab:human_control_answers}.%
\footnote{
Notice that for all our numerical analyses, the additional option (``I don't know'') was filtered out.
As shown in Tables~\ref{tab:acceptance_overal_results} and~\ref{tab:human_control_overal_results}, only very few people have chosen this answer option (1.89\% wave 1 and 0.67\% in wave 2).
}

\begin{table*}[thb]
\centering
\caption{Detailed descriptions of the considered scenarios.}
\label{tab:scenarios}
        \begin{tabular}{p{4.2cm} p{8.7cm}}
        \toprule
        \textbf{Scenario} & \textbf{Description} (with relevant references provided in brackets) \\
        \midrule
        \textbf{Fake news detection} & Identification of fake news in a social network. \citep{choudhary2021review,Khanam2021FakeNews,Hakak2021Ensemble} \\ \rowcolor{LightGray}
        \textbf{Insurance premium} & Determining the premium for liability insurance based on an individual's claims history. \citep{loi2021choosing,Cevolini2020insurance,baumann2023fairnessrisk} \\
        \textbf{Loan decision} & Deciding whether a person should get a loan based on their credit history. \citep{FUSTER2022Predictably,Sadok2022AIBank,KOZODOI2022CreditScoring} \\ \rowcolor{LightGray}
        \textbf{Medical diagnosis} & Making a diagnosis on the basis of a patient's medical data. \citep{Davenport2019AIhealthcare,rajpurkar2022ai,Williams2024MedicalDiagnosis,Yuri2021promise,Sivaraman2023} \\
        \textbf{Hiring} & Selecting a person for an interview based on their CV. \citep{harwell2022face,Booth2021Bias,kochling2020discriminated,Wilson2021ScreeningAudit,bogen2018help} \\ \rowcolor{LightGray}
        \textbf{Therapy discontinuation} & Decision to discontinue therapy based on a patient's medical condition. \citep{Davenport2019AIhealthcare,Labinsky2023ClinicalDecision} \\
        \textbf{Prisoner release} & Deciding whether to release an inmate based on various data (criminal history, behavior in prison, etc.). \citep{angwin2016machine,blomberg2010validation,berk2017impact,Travaini2022MLCriminalJustice} \\
        \bottomrule
        \end{tabular}
\end{table*}

\begin{table*}[thb]
\centering
\caption{
Answer options for different levels of human control over decisions made.
The two answer options \textit{AI decision with human oversight} and \textit{AI-assisted human decision} both fall into the category of {\color{humanAIcollaboration} \textbf{human-AI collaboration}}.
}
\label{tab:human_control_answers}
\begin{tabular}{p{3.3cm} p{10cm}}
\toprule
\textbf{Answer option} & \textbf{Description} \\
\midrule
{\color{AIonly} \textbf{AI only}} & {\color{AIonly} An AI system can decide this independently and automatically.} \\  \rowcolor{LightGray}
{\color{humanAIcollaboration} \textbf{AI decision with human oversight}} & {\color{humanAIcollaboration} An AI system can make this decision itself, but a human should always check the decision and intervene if necessary.} \\
{\color{humanAIcollaboration} \textbf{AI-assisted human decision}} & {\color{humanAIcollaboration} A human should make the decision, but an AI system should give advice on what the optimal decision might be.} \\ \rowcolor{LightGray}
{\color{humanOnly} \textbf{Human only}} & {\color{humanOnly} Only a human should make such decisions, without the involvement of AI.} \\
{\color{IDontKnow} \textbf{I don't know}} & {\color{IDontKnow} I do not know.} \\
\bottomrule
\end{tabular}
\end{table*}

\subsection{Sample description}
\label{ssection:Sample_description}

First, we provide an overview of the survey respondents across both waves of the study.
We provide more details in Figure~\ref{fig:descriptive_stats} in Appendix~\ref{app:Sample_description}.

\paragraph{Gender.}
Regarding gender distribution, in wave 1, 740 respondents identified as men, 769 as women, 2 as other, and 3 chose not to answer.
In wave 2, the numbers slightly varied, with 739 men and 741 women, 4 identifying as other, and 4 opting not to disclose their gender.

\paragraph{Language.}
Only a few participants are from the Italian-speaking region, which is also the language region with fewer inhabitants compared to the German- and French-speaking regions.

\paragraph{Age.}
The age distribution of respondents remained consistent across survey waves. In wave 1, the average age of participants was 49.73 years, while in wave 2, it was at 50.27 years.
For analysis, we categorized the participants into three age groups: under 40, 40-64, and 65 and above.

\paragraph{Education.}

We used the following ten standard Swiss education levels:
(1) Primary school or no degree,
(2) compulsory school,
(3) transitional education,
(4) general education without a baccalaureate,
(5) basic vocational training or apprenticeship,
(6) baccalaureate or similar (high school, vocational baccalaureate, specialized baccalaureate),
(7) post-secondary non-tertiary level (additional or secondary education),
(8) higher vocational training with federal certificate,
(9) university of applied sciences, university (including bachelor's, master's, and postgraduate studies),
(10) doctorate, habilitation,
(11) other,
(12) no answer.
For the analysis, we grouped education levels into three categories: university degree (levels 9-10), advanced (levels 6-8), and general (levels 1-5 \& 11-12).

In both waves, the majority of respondents reported having a university degree or higher (levels 9-10), with 740 individuals in wave 1 and 702 in wave 2.
About 30\% of participants of both survey waves had an advanced degree (levels 6-8).
Only 318 (350 in wave 2) participants fell in the general education bucket.

\paragraph{Political orientation.}
In terms of political orientation, respondents in wave 1 reported an average score of 4.14, whereas wave 2 respondents reported an average of 4.36 on a scale from 0 (far left) to 10 (far right).

\paragraph{ChatGPT awareness.}
The second wave of our survey indicates a high level of awareness about ChatGPT among participants, as shown in Table~\ref{tab:ChatGPT_awareness}.
The results reveal that a vast majority (96.37\%) are aware of this new technology.
Specifically, less than 4\% of participants have never heard of ChatGPT, and almost half of the participants have even used it themselves.
These numbers underscore the widespread awareness of AI among the public.
\begin{table*}[ht]
\caption{Survey responses on ChatGPT awareness and usage.}
\label{tab:ChatGPT_awareness}
\centering
\begin{tabular}{lc}
\toprule
\textbf{Answer option} & \textbf{Count (in \%)} \\

\midrule
I have never heard of a program called ChatGPT & 54 (3.6\%) \\ \rowcolor{LightGray}
I have heard of ChatGPT but have not looked into it & 287 (19.3\%) \\
I have read about ChatGPT but have never tried it myself & 479 (32.2\%) \\ \rowcolor{LightGray}
I have tried ChatGPT myself, but don't use it otherwise & 448 (30.1\%) \\
I use ChatGPT regularly for private and/or professional purposes & 220 (14.8\%) \\
\bottomrule
\end{tabular}
\end{table*}
\\\\
For the remainder of this manuscript, all results are provided based on weighted survey data.

\subsection{Data analysis}
\label{ssec:Survey_quotas_and_weights}

We used two mechanisms to ensure representativeness.
First, we set quotas of age, gender, and language region when collecting the data to make sure we have a balanced sample according to these demographic characteristics.
Second, for both surveys, we weighted the data of the final sample to ensure representativeness with respect to age, gender, language region, education, and political orientation.
The weighting was done by the professional survey company and is based on standard survey weighting data from the Swiss Population Census~\citep{bfs2022}.
We used the open-source \texttt{survey} package~\citep{Lumley2024survey} for the survey-weighted analyses in R.

Unless stated otherwise, we present weighted means with 95\% CI in the following analysis.
We report adjusted p-values~\citep{Yekutieli1999}, following the Benjamini–Hochberg procedure to control the false discovery rate for multiple hypothesis testing~\citep{Benjamini1995}.
We denote statistical significance with *** if p-value $< 0.001$, ** if p-value $< 0.01$, * if p-value $< 0.05$, and not significant (ns) otherwise.

\section{Results}
\begin{figure*}[h!]
\centering
\includegraphics[width=0.82\textwidth]{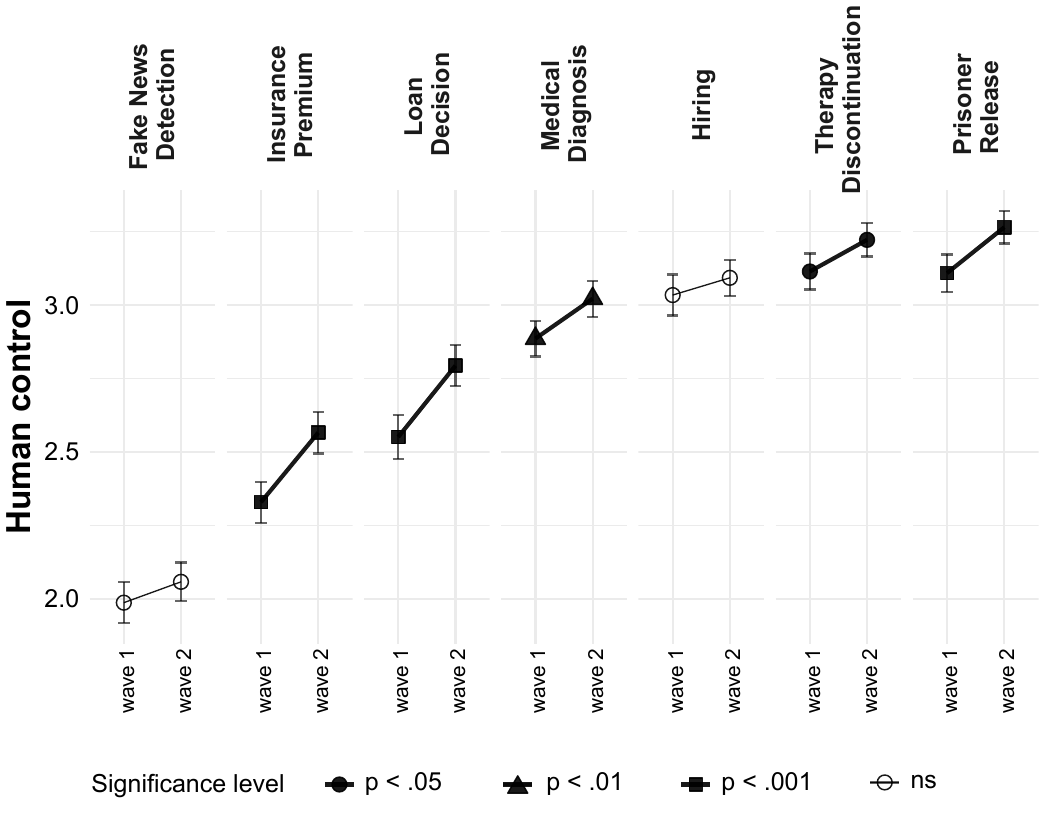}
\caption{
The generative AI boom is associated with a significant increase in required human control for five out of seven scenarios.
In both survey waves, human control preferences are heavily dependent on context.
Just as in Figure~\ref{fig:average_acceptance_comparison_by_scenario}, we show weighted means with 95\% CI and significance levels indicate statistically significant differences across survey waves as determined by a two-sample design-based t-test.
}
\label{fig:human_control_average_comparison_by_scenario}
\end{figure*}

\begin{tcolorbox}[colframe=gray!90, colback=gray!20, title=Summary of the main findings, fonttitle=\bfseries, boxrule=0.5pt, left=5pt, right=5pt, top=5pt, bottom=5pt, boxsep=5pt]
\begin{enumerate}[left=0pt, topsep=2pt, itemsep=3pt, parsep=0pt, label=RQ\arabic*:]
    \item \textbf{GenAI boom:} The recent surge in generative AI is associated with a \textbf{decrease in overall AI acceptance} and an \textbf{increased demand for human oversight}.
    The associations are statistically significant for almost all scenarios (see Figures~\ref{fig:average_acceptance_comparison_by_scenario} and~\ref{fig:human_control_average_comparison_by_scenario}).
    Yet, AI acceptance and human control preferences remain highly contextual across both waves.
    \item \textbf{Amplification of existing inequalities:} Our study reveals that the \textbf{GenAI boom is associated with exacerbated pre-existing disparities} in AI acceptance:
    \begin{itemize}
        \item Respondents with university degrees show consistently higher acceptance of AI compared to those with advanced non-university education, with the gap widening after the GenAI boom.
        \item French-speaking respondents demonstrate lower AI acceptance than German speakers, a difference that remains stable across most scenarios but has increased for medical contexts since the GenAI boom.
        \item Women express significantly more skepticism toward AI in medical scenarios (medical diagnosis and therapy discontinuation), and this gender gap has grown further post-GenAI boom.
    \end{itemize}
\end{enumerate}
\end{tcolorbox}

\subsection{RQ1a: Reduced AI acceptance with the GenAI boom}

The average acceptance of AI was significantly lower at wave 2 compared to wave 1 across five out of seven examined scenarios post-GenAI boom, as depicted in Figure~\ref{fig:average_acceptance_comparison_by_scenario}. Table~\ref{tab:acceptance_overal_results} shows the response distribution for each wave, irrespective of specific scenarios, in more detail.
Notably, the fraction of respondents who found AI ``not acceptable at all'' increased from 23.87\% in wave 1 to 30.09\% in wave 2, indicating a stark rise in strong opposition to AI applications.
Conversely, the percentage of people deeming AI ``definitely acceptable'' decreased from 13.07\% to 10.47\%.

\begin{figure*}[h!]
    \centering
    \includegraphics[width=1\textwidth]{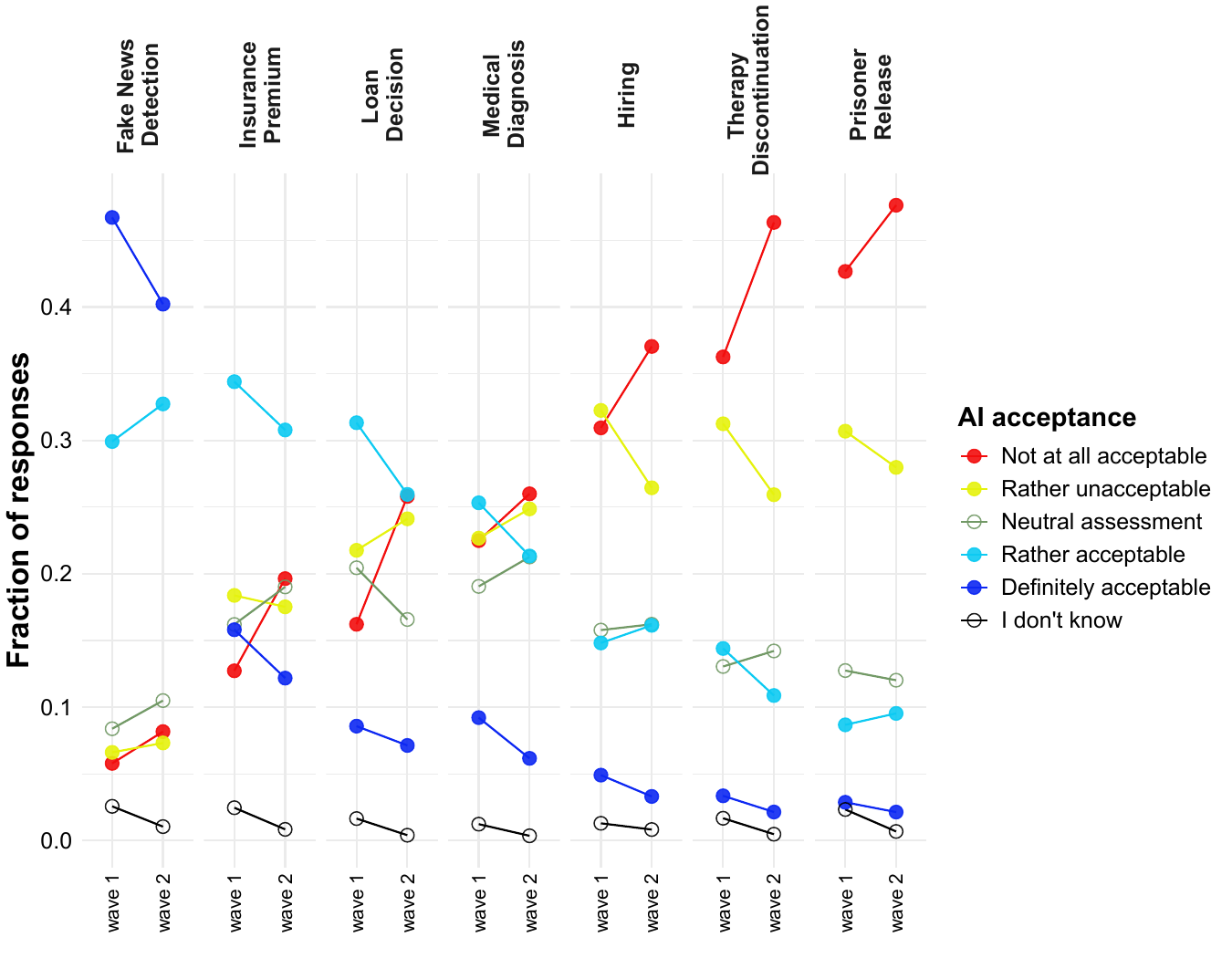}
    \caption{Response distribution in \% for acceptance questions across survey waves and scenarios.}
    \label{fig:acceptance_fractions_comparison_by_scenario}
\end{figure*}

Scenarios involving high-impact decisions, such as prisoner release, therapy discontinuation, and hiring decisions, showed the lowest acceptance levels in both waves.
But even scenarios with much higher acceptance levels, such as fake news detection and insurance premiums, showed pronounced increases in people finding the use of AI not acceptable at all.
With the generative AI boom, all scenarios exhibited consistent declines in the share of people perceiving the use of AI as ``definitely acceptable''.
These effects are consistent across all scenarios, as can be seen in Figure~\ref{fig:acceptance_fractions_comparison_by_scenario}.

These results suggest a growing skepticism towards AI in critical decision-making areas, highlighting increased concerns about the reliability and ethics of AI technologies, especially in sensitive contexts.

\begin{table}[h!]
\caption{
Overview of responses regarding the perceived necessity for human control. Values show weighted averages across all scenarios. The survey-weighted chi-square test for overall distribution change between the two waves is statistically significant ($p < 0.001$). The significance column shows results from individual chi-square tests for each response option, indicating whether the proportion of that response changed significantly between waves.
}
\label{tab:acceptance_overal_results}
\centering
\begin{tabular}{lccc}
\toprule
\textbf{Response type} & \textbf{Wave 1} & \textbf{Wave 2} & \textbf{Significance} \\
\midrule
Not at all acceptable & 23.87\% & 30.09\% & ** \\
Rather unacceptable & 23.38\% & 22.03\% & ns \\
Neutral assessment & 15.10\% & 15.69\% & ns \\
Rather acceptable & 22.70\% & 21.05\% & ns \\
Definitely acceptable & 13.07\% & 10.47\% & * \\
I don't know & 1.89\% & 0.67\% & * \\
\bottomrule
\end{tabular}
\end{table}

\subsection{RQ1b: Increased need for human control with the GenAI boom}
\label{sssec:Increased_need_for_human_control}

\begin{figure*}[h!]
    \centering
    \includegraphics[width=1\textwidth]{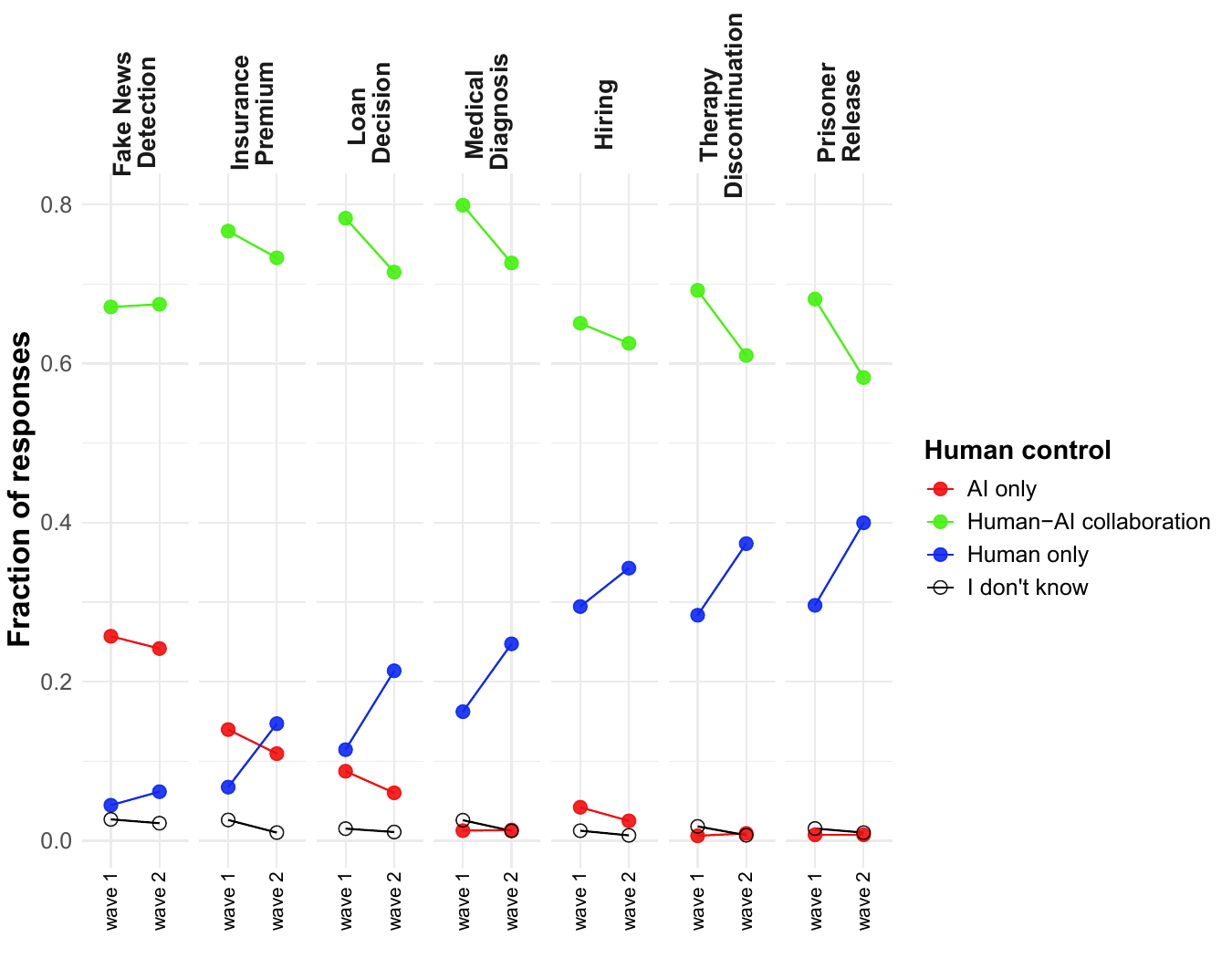}
    \caption{Response distribution in \% for human control questions across survey waves and scenarios.
    Notice that in this figure, the two answer options (\textit{AI decision with human oversight} and \textit{AI-assisted human decision}) are merged into one single category titled \textit{human-AI collaboration}, as they are conceptually very similar -- see Table~\ref{tab:human_control_answers} for more details.
    In Appendix~\ref{app:human_control_without_grouped_answer}, we provide the same overview but consider both answer options separately, showing that the overall results remain unaffected.
    However, notice that the overall reduction of the share of people favoring human-AI collaboration is mainly driven by fewer respondents opting for AI decisions with human oversight.
    }
    \label{fig:human_control_fractions_simple_comparison_by_scenario}
\end{figure*}

The reported necessity for human oversight in AI decisions increased significantly across almost all scenarios, as illustrated in Figure~\ref{fig:human_control_average_comparison_by_scenario}.
Table~\ref{tab:human_control_overal_results} shows the response distribution in more detail, and Figure~\ref{fig:human_control_fractions_simple_comparison_by_scenario} shows the results across the different scenarios.
Interestingly, people predominantly favor human-AI collaboration (human in or on the loop) across all scenarios, namely, 72.06\% of respondents in wave 1 and 66.68\% in wave 2.
A considerable fraction of people think AI should not be involved in the decision process at all, 18.04\% before and 25.52\% after the generative AI boom.

Scenarios with high-impact decisions, such as prisoner release, therapy discontinuation, hiring, or medical diagnosis, showed the highest desire for human control levels across both waves, with about 40\% of people requiring only humans to be involved in wave 2.
Other scenarios have also seen substantial increases in the reported necessity for human oversight:
For example, the average reported human control values rose from 2.33 to 2.58 in the insurance premium scenario and from 2.55 to 2.81 in the banking scenario related to granting loans.

In a similar way to AI acceptance levels, this trend highlights a rising concern about fully automated decision-making in sensitive contexts, and people want humans to be in the loop more often.

\begin{table}[h!]
\caption{
Overview of responses regarding the perceived necessity for human control. Values show weighted averages across all scenarios. The survey-weighted chi-square test for overall distribution change between the two waves is statistically significant ($p < 0.001$). The significance column shows results from individual chi-square tests for each response option, indicating whether the proportion of that response changed significantly between waves.
}
\label{tab:human_control_overal_results}
\centering
\begin{tabular}{lccc}
\toprule
\textbf{Response type} & \textbf{Wave 1} & \textbf{Wave 2} & \textbf{Significance} \\
\midrule
AI only & 7.90\% & 6.67\% & ns \\
AI decision with human oversight & 29.98\% & 26.07\% & ** \\
Human decision with AI assistance & 42.08\% & 40.61\% & ns \\
Human only & 18.04\% & 25.52\% & *** \\
I don't know & 2.00\% & 1.14\% & ns \\
\bottomrule
\end{tabular}
\end{table}

\subsection{RQ2: Amplification of existing inequalities with the GenAI boom}

\begin{table}[!htbp] \centering 
  \caption{Survey-weighted linear regression results for mean AI acceptance and human control across all scenarios. Coefficients that are statistically significant are highlighted in bold.}
  \label{tab:regression}
\begin{tabular}{@{\extracolsep{5pt}}lcccc} 
\toprule
& \multicolumn{2}{c}{\textbf{AI acceptance}} & \multicolumn{2}{c}{\textbf{Human control}} \\ 
\cline{2-3} \cline{4-5} 
 & Wave 1 & Wave 2 & Wave 1 & Wave 2 \\ 
 Gender (f) & $-$0.037 & $-$0.111 & 0.037 & $-$0.005 \\ 
  & (0.078) & (0.075) & (0.047) & (0.043) \\ 
  General education & $-$0.103 & \textbf{$-$0.155$^{*}$} & 0.047 & 0.055 \\ 
  & (0.082) & (0.073) & (0.050) & (0.045) \\ 
  University degree & 0.042 & \textbf{0.148$^{*}$} & $-$0.080 & \textbf{$-$0.111$^{**}$} \\ 
  & (0.067) & (0.058) & (0.041) & (0.037) \\ 
  German language & 0.017 & 0.147 & 0.065 & \textbf{$-$0.239$^{*}$} \\ 
  & (0.188) & (0.193) & (0.129) & (0.105) \\ 
  French language & $-$0.121 & $-$0.221 & 0.168 & 0.004 \\ 
  & (0.200) & (0.197) & (0.134) & (0.110) \\ 
  Digital literacy & 0.053 & 0.070 & $-$0.035 & $-$0.044 \\ 
  & (0.047) & (0.040) & (0.030) & (0.025) \\ 
  Age (40-64) & \textbf{$-$0.225$^{*}$} & $-$0.048 & \textbf{0.136$^{**}$} & \textbf{0.104$^{*}$} \\ 
  & (0.095) & (0.074) & (0.053) & (0.053) \\ 
  Age (65+) & 0.075 & 0.135 & 0.022 & 0.037 \\ 
  & (0.097) & (0.114) & (0.052) & (0.058) \\ 
  Political orientation & 0.020 & 0.012 & 0.007 & $-$0.001 \\ 
  & (0.014) & (0.015) & (0.010) & (0.009) \\ 
  Constant & \textbf{2.673$^{***}$} & \textbf{2.380$^{***}$} & \textbf{2.619$^{***}$} & \textbf{3.112$^{***}$} \\ 
  & (0.264) & (0.239) & (0.180) & (0.146) \\ 
\bottomrule
\textit{Note:}  & \multicolumn{4}{r}{$^{*}p<0.05$; $^{**}p<0.01$; $^{***}p<0.001$} \\ 
\end{tabular} 
\end{table}

The survey-weighted linear regression analysis in Table~\ref{tab:regression} reveals several demographic predictors of AI acceptance and preferred human control (across all scenarios), highlighting nuanced trends.
The level of education plays a key role in wave 2, with university-educated respondents showing a significant positive association with AI acceptance (compared to people with advanced education) and a tendency toward reduced emphasis on human control, suggesting higher confidence in AI autonomy.
On the other hand, general education is associated with lower acceptance in wave 2.
Age (40-64) also emerges as a significant predictor: it shows that older people (40-64) find AI use generally less acceptable than young people (<40) pre-GenAI boom.
The significant positive predictor of human control in wave 1 for the same age group mirrors this result.
Furthermore, German language is negatively associated with human control post-GenAI boom
Meanwhile, when looking at the averages across all scenarios, we do not find any significant effects for digital literacy, political orientation, and gender.

The survey results further indicate that the generative AI boom is associated with the reinforcement of existing group disparities for language regions and gender, but only in specific scenarios.
In the following, we summarize the amplification of these disparities between wave 1 and 2.
Please refer to Appendix~\ref{ssec:RQ3_continued} for the associated figures and additional details.

\subsubsection{Education gap.}

The GenAI boom is associated with the widening of the existing education-based disparities in AI acceptance (see Figure~\ref{fig:education_gap}).
Respondents with higher levels of education consistently exhibit greater acceptance of AI.
Our findings show a clear trend: individuals with university degrees demonstrate the highest levels of AI acceptance, followed by those with advanced non-university education, while those with only general education express the lowest acceptance levels.
This hierarchy is evident across all surveyed scenarios.

Crucially, the gap between these educational groups has further expanded during the the GenAI boom.
While university-educated respondents maintained relatively stable acceptance levels, those with lower levels of education became increasingly critical of AI use.
For more details, see Appendix~\ref{sssec:Amplified_education_gap}.

\subsubsection{Language region gap.}
Our results show that exacerbated language-based disparities in AI acceptance have also been associated with the GenAI boom (see Figure~\ref{fig:language_gap}).
French-speaking respondents demonstrate lower AI acceptance compared to German-speaking respondents, and their acceptance significantly decreased across the two survey waves in medical contexts.
For more details, see Appendix~\ref{sssec:Amplified_language_gap}.

\subsubsection{Gender gap in health scenarios.}
The GenAI boom is also associated with the deepening of the gender gap in AI acceptance, especially in health-related scenarios (see Figure~\ref{fig:gender_gap}).
Additionally, while women's AI acceptability has significantly decreased in five out of seven scenarios, for men, this effect was only significant in the loan decision scenario (see top panel of Figure~\ref{fig:gender_gap}).
Women express significantly more skepticism toward the use of AI for medical diagnosis and therapy discontinuation compared to men, and this gender gap has widened further post-GenAI boom.
For more details, see Appendix~\ref{sssec:Amplified_gender_gap}.

\section{Discussion}

Our study highlights the critical role of contextuality in public acceptance of AI technologies, particularly in light of the recent generative AI boom.
In high-impact scenarios like therapy discontinuation, where outcomes could be life-threatening, only about 3\% of respondents view AI use without human oversight as acceptable.
In contrast, in scenarios such as fake news detection, around 40\% of respondents find AI use entirely acceptable. 
In contrast to~\citet{kaufmann_task-specific_2023,jussupow_why_2020}, we find that not only medical scenarios but also hiring decisions and prisoner release decisions exhibit consistently low AI acceptances.
Our observed skeptical tendency towards AI tells a very different story than a similar survey conducted in Britain, which found that people's perceived benefit levels outweigh concerns regarding AI use~\citep{modhvadia2023people}.
However, there are two key differences between these two studies:
First, while they consider very general AI uses (including use cases like virtual AI assistants or AI-based simulations for advancing knowledge), we focus specifically on high-impact decision-making scenarios.
Second, they ran their study before the GenAI boom, which has ``already
impacted public discourse towards some AI technologies since [their] survey''~\citep[p.~57]{modhvadia2023people}.
Furthermore, our findings of varying acceptance in high-impact scenarios align with~\citet{castelo_task-dependent_2019} observation that people trust algorithms less in more subjective tasks.
However, while they studied algorithm trust before the GenAI boom using a simplified definition of algorithms, our study examined AI acceptance after this technological shift with more detailed AI priming, potentially explaining some of the differences in overall sentiment.

In the following, we address several key aspects of our findings. We begin by examining how various AI definitions -- including our own approach -- shape research outcomes and policy considerations. Next, we explore the implications of public opinion shifts following the generative AI boom and contextualize the socio-demographic variations in AI acceptance we observed against the more general patterns of digital inequalities in Switzerland. We then consider how these changing public perceptions and socio-demographic differences could inform AI governance and regulatory frameworks. Finally, we acknowledge our study's limitations and propose future research directions that could bridge our survey-based methodology with the experimental designs more commonly employed in human-computer interaction (HCI) research on AI acceptance and trust.

\subsection{The implications of AI definitions for research and policy}
Our study's findings must be interpreted considering how ``artificial intelligence'' was defined for participants. This reflects a broader challenge in both research and policy: how different conceptions of AI influence public attitudes and regulatory responses.

Public and scholarly discourse around AI encompasses multiple definitions ranging from narrow task-specific systems to hypothetical general intelligence~\cite{narayanan_ai_2024}. Our study focused on actual AI applications in use today, yet public attitudes may be shaped by broader cultural narratives about AI from science fiction and media~\cite{narayanan_ai_2024,cave_ai_2020}. These definitional ambiguities create challenges for interpreting public opinion and developing governance frameworks.

First, public concerns may target different phenomena than those addressed by current policy initiatives. Despite our priming that aimed to focus participants on contemporary existing AI systems, survey respondents expressing concern about AI-based decision-making may be imagining more autonomous general intelligence rather than the statistical models currently widely deployed. The significant shift in AI acceptance we observed following the generative AI boom may reflect not just reactions to new technologies, but evolving conceptions of what constitutes ``AI'' in public understanding. Second, technical experts, policymakers, media commentators, and the general public often use the term ``AI'' differently~\cite{krafft_defining_2020,narayanan_ai_2024}, potentially leading to misaligned expectations between these groups. This definitional fragmentation may partially explain the socio-demographic differences we observed -- for example, educational background or gender may correlate with exposure to particular framings of AI~\cite{bao_whose_2022,cave_ai_2020}.

These definitional challenges have several implications for future research and policy development. Research on public attitudes toward AI should explicitly acknowledge how definitional choices may influence findings. Studies might systematically vary AI definitions provided to participants to understand how different framings affect reported attitudes. Such research could help disentangle concerns about current applications from fears about hypothetical future capabilities. For policymakers, these challenges suggest the importance of granular, technology-specific approaches rather than broad regulations of ``AI'' as a monolithic category. Our findings on context-dependent acceptance patterns provide empirical support for such granularity.

\subsection{Generative AI boom and public opinion}
One of the key findings of our study is the decline in overall AI acceptance and the increased demand for human control following the generative AI boom.
Despite the growing prevalence of AI in everyday applications, respondents showed heightened concerns over AI use, particularly in high-impact contexts such as medical diagnosis and therapy discontinuation.
The generative AI boom appears to have raised awareness about the limitations and risks of autonomous decision-making, thereby prompting calls for stricter human control and oversight, which is in line with the expectations that can be derived from related work~\cite{Sundar2020,Babamiri2022Insights,mays2022ai}.

Media coverage and public discourse around AI have likely played a major role in shaping these shifting attitudes. The period between our two surveys (2022-2023) saw extensive media reporting on AI in Switzerland, as we show in Figure~\ref{fig:aiboom}. News stories highlighting hallucinations, biases, and other failures of large language models likely contributed to the increased skepticism we observed, as media framing of AI can significantly influence public perceptions~\citep{cui_influence_2021,scheufele_public_2005,ho_trust_2024}. We suggest that the differential shifts in AI acceptance between Swiss linguistic regions can be partially attributed to differences in how AI was framed across these information environments. While our observations on these differential shifts may seem particularly relevant to officially multilingual countries like Switzerland or Belgium, they also apply to officially monolingual contexts, such as the United States, where a significant portion of the population communicates primarily in languages other than English, such as Spanish. Future work should systematically analyze such media coverage to better understand these mechanisms.

\subsection{Socio-demographic differences in AI acceptance}

The surge in interest and usage of AI technologies, exemplified by ChatGPT, has not only altered public perceptions generally but also introduced specific challenges related to the potential reinforcement of existing social inequalities.
The educational divide in AI acceptance suggests potential knowledge or exposure gaps that could exacerbate existing digital inequalities. Those with higher education levels demonstrated greater acceptance of AI in decision-making contexts, potentially reflecting greater familiarity with technological systems or higher perceptions of AI usefulness for decision-making -- a tendency previously found by~\citet{araujo_ai_2020}.
Our findings suggest that educational disparities may create uneven capabilities to understand, evaluate, and potentially benefit from AI systems.
This observation mirrors broader patterns of digital inequality in Switzerland, where education strongly predicts internet use, skills and privacy-protective behaviors~\cite{buchi_chapter_2021,kappeler_left_2021,festic_its_2021}.
\citet{buchi_chapter_2021} found that lower education is associated with decreased ability to protect one's privacy online, and~\citet{kappeler_left_2021} showed that lower-educated people are less likely to be internet users, with this gap persisting between 2011 and 2019.

While a gender gap in AI acceptance has been reported by existing studies~\citep{mendez-suarez_factors_nodate,araujo_ai_2020,sartori_minding_2023,cho_dual_2024,horowitz_what_2021}, we find that it increased further with the GenAI boom, particularly in health-related contexts. This pattern may reflect gendered experiences with technology, different risk assessments with regard to technology~\citep{araujo_ai_2020}, or responses to the documented gender biases in many AI systems~\citep{oconnor_gender_2024}. These gender differences also align with findings on digital inequality in Switzerland more broadly~\cite{buchi_chapter_2021,kappeler_left_2021,festic_its_2021}. Future research should investigate the underlying mechanisms of these disparities and identify ways to mitigate them, focusing on why certain demographic groups exhibit disparate preferences when it comes to AI use in high-impact scenarios.

The demographic disparities in AI acceptance we observed align with existing digital divides in Switzerland~\cite{buchi_chapter_2021,festic_its_2021,kappeler_left_2021}, suggesting that AI technologies may become another domain where digital inequalities are manifested along the lines of gender and education. The widening of these gaps following the generative AI boom is particularly concerning, as it indicates that rapid technological advancement may exacerbate rather than ameliorate existing inequalities. As AI systems become more integrated into essential services and opportunities, these acceptance disparities could translate into uneven adoption and benefit, further disadvantaging already marginalized groups. Addressing these emerging AI divides requires coordinated efforts from both researchers and policymakers: researchers should investigate mechanisms underlying these disparities and develop inclusive design approaches, while policymakers should consider educational interventions and regulatory frameworks that ensure equitable access to AI literacy and benefits across demographic groups.

\subsection{Implications for AI governance and regulation}
Our findings on shifting public attitudes following the generative AI boom also highlight a policy-related challenge: how to incorporate public opinion into governance frameworks for rapidly evolving technologies~\cite{marchant_growing_2011}. The significant changes in AI acceptance we observed between 2022 and 2023 demonstrate how quickly public sentiment can evolve in response to technological developments, raising important questions about the appropriate role of public opinion in AI governance. This timing challenge is further complicated by the fact that regulation typically takes years to develop and implement. For example, the EU AI Act was initially proposed in 2021 but only finalized in 2024~\cite{noauthor_regulation_2024}, during which time public opinion and AI capabilities both evolved significantly. 

These challenges suggest the need for more adaptive regulatory frameworks that can evolve with public opinion and technological capabilities. While the specifics of such frameworks should be discussed in-depth by legal and policy scholars, based on relevant research we anticipate the following approaches to be particularly useful for the incorporation of public opinion into AI governance and regulation. First, one could create participatory governance structures that would establish formal roles for diverse public stakeholders in ongoing oversight of AI systems, potentially through citizen advisory boards or similar mechanisms, as discussed by~\cite{cihon_fragmentation_2020,bullock_aligning_2022}. Second, regulatory sandboxes providing controlled environments where AI applications can be tested with public input before wider deployment could be helpful -- similar to approaches used in fintech regulation. See~\cite{oecd_regulatory_2023} for a detailed discussion of this approach in the context of AI.

Another key insight from our study is that public acceptance varies significantly by context, even among applications that might all be categorized as ``high-risk'' under frameworks like the EU AI Act~\cite{noauthor_regulation_2024}. This suggests that regulatory approaches that differentiate more finely between contexts might better align with public expectations. For example, while the EU AI Act categorizes both healthcare AI and HR AI systems as ``high-risk,'' our findings show that public acceptance differs between medical diagnosis (with higher acceptance) and hiring decisions (with lower acceptance). These nuances could inform more contextually appropriate requirements for human oversight, explainability, and other safeguards.

Additionally, our findings on socio-demographic differences in AI acceptance highlight potential inequities in how AI governance may be perceived. The growing educational divide and gender gap in AI acceptance suggest that regulatory approaches may need to specifically address the concerns of groups showing lower acceptance to build broader public trust.

\subsection{Limitations and future work}
\label{ssec:Limitations_and_future_work}

As with any research, our work is not without limitations that offer important opportunities for future research.
While our study highlights issues related to context-dependent AI acceptance and the amplification of socio-demographic differences, it does not clarify the causal mechanisms behind these trends.
Future research should investigate these underlying mechanisms further, offering policymakers and industry professionals actionable insights to promote responsible AI adoption and equitable outcomes.

Due to the constraints on questionnaire length, we could not fully describe the nuances of the different high-impact scenarios.
Consequently, we cannot be certain that our priming of the AI definition worked so that all participants formed equivalent mental models of AI-based decisions~\citep{Kulesza2012,Bansal2019,Cabrera2023}. Elements such as AI prediction performance and its influence on human decisions may also play a role~\cite{Bussone2015Systems,kocielnik_will_2019,jacobs2021machine}, along with the types of biases present~\citep{Rastogi2022,Baumann2023biasondemand}, especially in high-impact contexts.
Given these limitations, our survey-based study is subject to classic issues like social desirability bias, as it does not involve behavioral experiments. 

An additional limitation with regards to the AI primer is that our study employed a specific operational definition of AI that emphasized current applications and explicitly distanced itself from more speculative or futuristic conceptions often portrayed in popular media. Public conceptions of AI are diverse and often include these more advanced or general AI systems~\citep{narayanan_ai_2024}. Our chosen framing may have influenced participants' responses by potentially narrowing their consideration to more mundane or familiar technologies, which could lead to different attitudes than if participants were considering more advanced AI capabilities. This specific framing may have additionally introduced or strengthened the effects of social desirability bias by implicitly suggesting that concerns about more advanced AI systems are ``unrealistic.'' Future work should consider comparing responses across different AI definitions or allowing participants to respond based on their own conception of AI.

Our methodology faces additional limitations regarding demographic data collection and literacy measurement. As demonstrated by~\citet{danaher2008stereotype}, the timing of demographic questions can induce stereotype threat effects that potentially influence results. Although we made deliberate choices to ensure consistency in our self-reported literacy measures between survey waves, these measures lack objective validation and may be subject to reporting biases. Similarly, our ChatGPT awareness assessment lacks comparative baselines against fictional technologies, making it difficult to distinguish between genuine knowledge and mere name recognition. Future research should employ validated measures with objective components and control for these methodological limitations.

Finally, even when based on representative samples, our findings may not be generalizable across countries due to the wide variation in AI acceptance across cultural and national contexts~\cite{kaufmann_task-specific_2023,fletcher_what_2024}.

\subsubsection{Methodological considerations for future research on AI acceptance}
Our study contributes to research on public perceptions of AI technologies but differs methodologically from much HCI literature, which often relies on experimental designs with specific user groups. This raises questions about how AI acceptance, trust, reliability, and other related constructs should be measured across research contexts.

While our study focused primarily on ``acceptance,'' measured through survey questions rather than experimental designs, the HCI literature has developed a rich ecosystem of related but distinct constructs that capture different aspects of how people perceive and interact with AI systems, measured through experiments. These include, for example, trust~\cite{vereschak_how_2021,klingbeil_trust_2024} or fairness~\cite{nakao_toward_2022,dodge_explaining_2019}. When it comes to AI acceptance specifically, experimental HCI designs primarily focus on the way different design factors such as various forms of explainability of AI decisions affect user acceptance of AI~\cite{ebermann_explainable_2023,maehigashi_experimental_2023,kocielnik_will_2019}.

These methodological differences reflect distinct research objectives. Our survey approach captures population-level attitudes to inform policy, while experimental HCI research isolates factors influencing user experience to improve design. This generalizability versus specificity tradeoff highlights their complementary nature.

Our context-dependent findings suggest opportunities to bridge these approaches. Experimental designs could investigate why acceptance varies across contexts by manipulating factors like decision importance and domain. For example, the lower acceptance in hiring versus medical contexts could be explored through experiments controlling for perceived reliability, fairness, and explainability.

The demographic differences we observed -- particularly in education and gender -- suggest experimental studies should use diverse samples and analyze interaction effects between demographic factors and design interventions. Such studies might examine whether different forms of explainability affect acceptance differently across demographic groups.

Both approaches face ecological validity challenges. Surveys rely on hypothetical scenarios that may not reflect actual behavior with AI systems, while experiments often use simplified tasks that do not capture real-world complexity. Field experiments deploying AI systems to diverse populations in natural settings could help address these limitations.

The rapid shift in AI acceptance following the generative AI boom highlights the temporal instability of public perceptions, suggesting both research traditions should incorporate longitudinal designs to capture evolving perceptions. By integrating insights from both approaches, researchers can develop more nuanced understandings of public AI perceptions and more effective system designs.

\section{Conclusion}

In this study, we leveraged a survey approach to capture public attitudes towards AI.
Our study design is unique as we conducted two survey waves, one before and one after the GenAI boom, revealing a decreased AI acceptance coupled with a pronounced preference for humans to be in/on the loop.
Our findings indicate a marked decline in public acceptance of AI, coupled with a heightened demand for human oversight, particularly in sensitive applications where the implications of decisions are profound. Notably, the GenAI boom is significantly associated with amplified existing disparities, with acceptance levels diverging along the lines of education, language, and gender.
These insights underscore the importance of context-sensitive AI developments and regulation frameworks ensuring appropriate levels of human oversight rather than adopting blanket policies.

\section*{Acknowledgements}

This work was supported by the National Research Programme ``Digital Transformation'' (NRP 77) of the Swiss National Science Foundation (SNSF), grant number 187473.

\bibliographystyle{plainnat}
\bibliography{references}

\appendix
\newpage

\section{Survey samples}
\label{app:Sample_description}

\begin{figure*}[htb]
    \centering
    \includegraphics[width=0.9\textwidth]{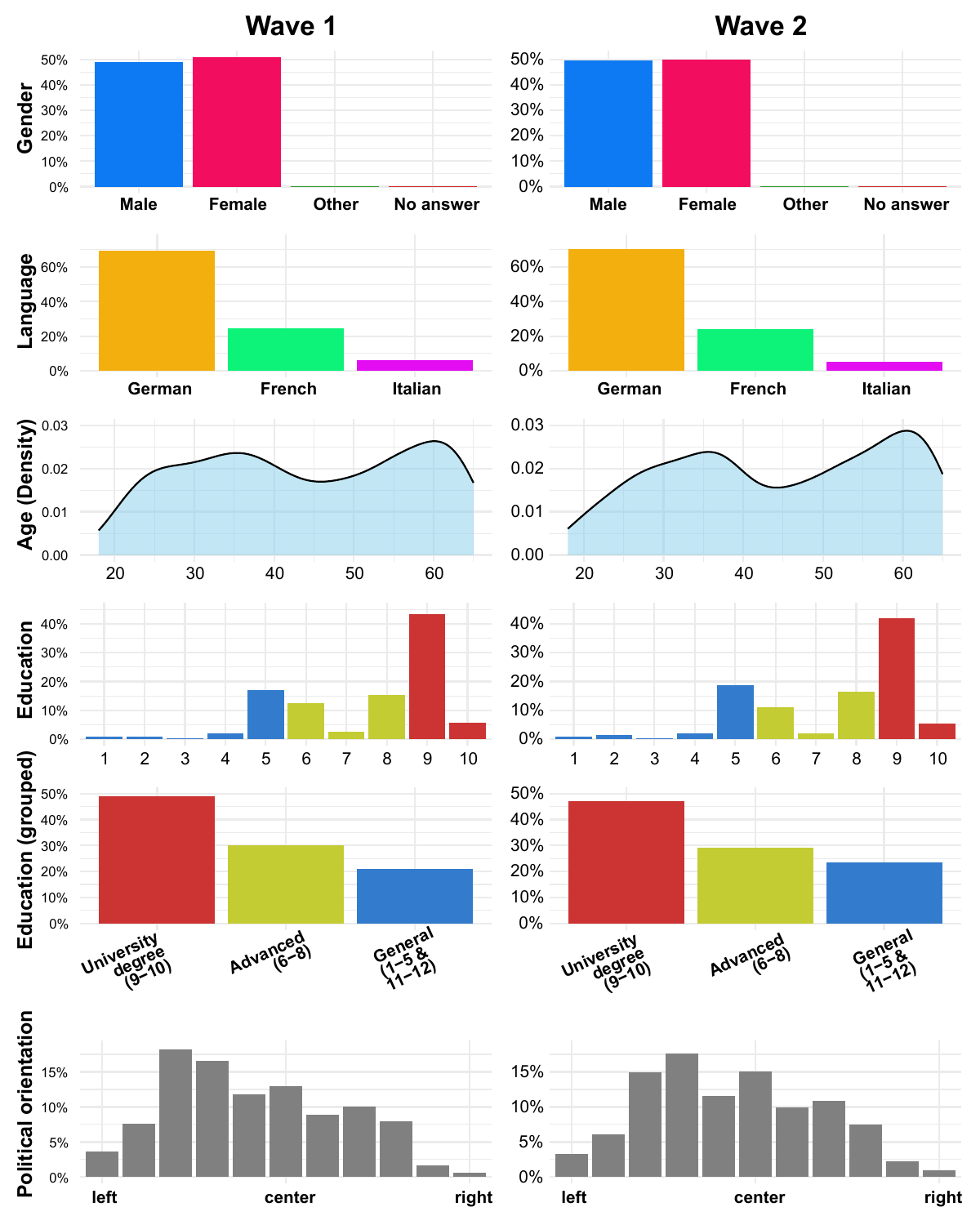}
    \caption{Demographic overview of survey respondents in both waves.}
    \label{fig:descriptive_stats}
\end{figure*}

Figure~\ref{fig:descriptive_stats} provides detailed information about the samples from both survey waves.

\section{Further empirical results}

\subsection{Perceived need for human control without aggregated answer options}
\label{app:human_control_without_grouped_answer}

\begin{figure*}[thb]
    \centering
    \includegraphics[width=1\textwidth]{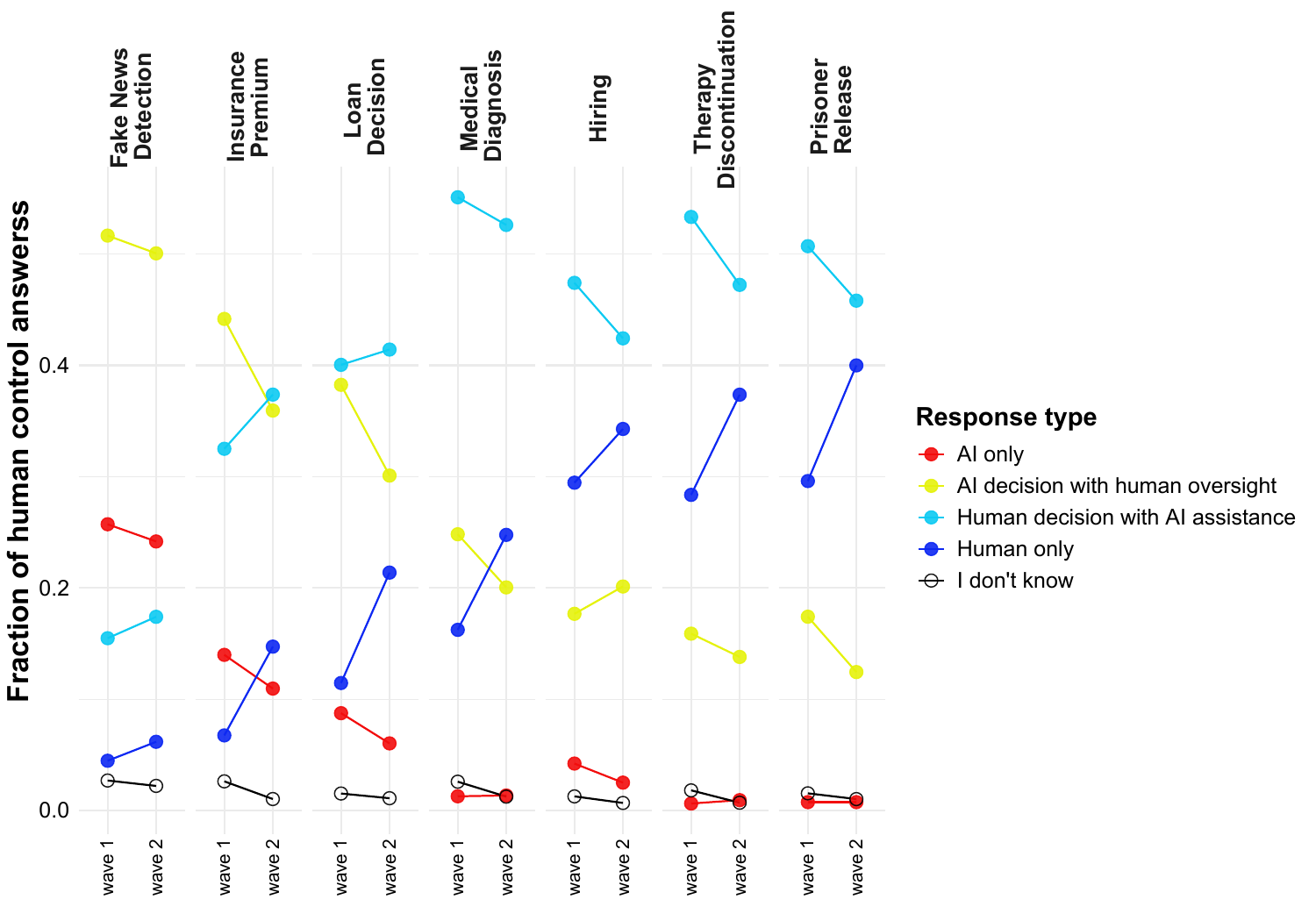}
    \caption{Response distribution in \% for human control questions across survey waves and scenarios.}
    \label{fig:human_control_fractions_comparison_by_scenario}
\end{figure*}

In Section~\ref{sssec:Increased_need_for_human_control}, we combined the two answer options (\textit{AI decision with human oversight} and \textit{AI-assisted human decision}) into one single category titled \textit{human-AI collaboration}, as they are conceptually very similar.
The resulting fraction of selected answer responses is shown in Figure~\ref{fig:human_control_fractions_simple_comparison_by_scenario}.
In Figure~\ref{fig:human_control_fractions_comparison_by_scenario}, we show the same overview but without aggregating any of the answer options.
The trend of more human oversight after the generative AI boom is clearly visible.
Much fewer people want AI to make the decision, with humans merely taking the role of intervening.
The only exception is the hiring scenario, where the fraction of people answering ``AI decision with human oversight'' slightly increased from 17.67 to 20.12.
The share of people favoring human decisions with AI assistance is more stable and even decreases with the generative AI boom for scenarios with higher potential impacts (medical diagnosis, hiring, therapy discontinuation, and prisoner release).

\subsection{Amplification of existing inequalities with the GenAI boom}
\label{ssec:RQ3_continued}

\subsubsection{Amplified education gap.}
\label{sssec:Amplified_education_gap}

\begin{figure*}[thb]
\centering
\includegraphics[width=0.9\textwidth]{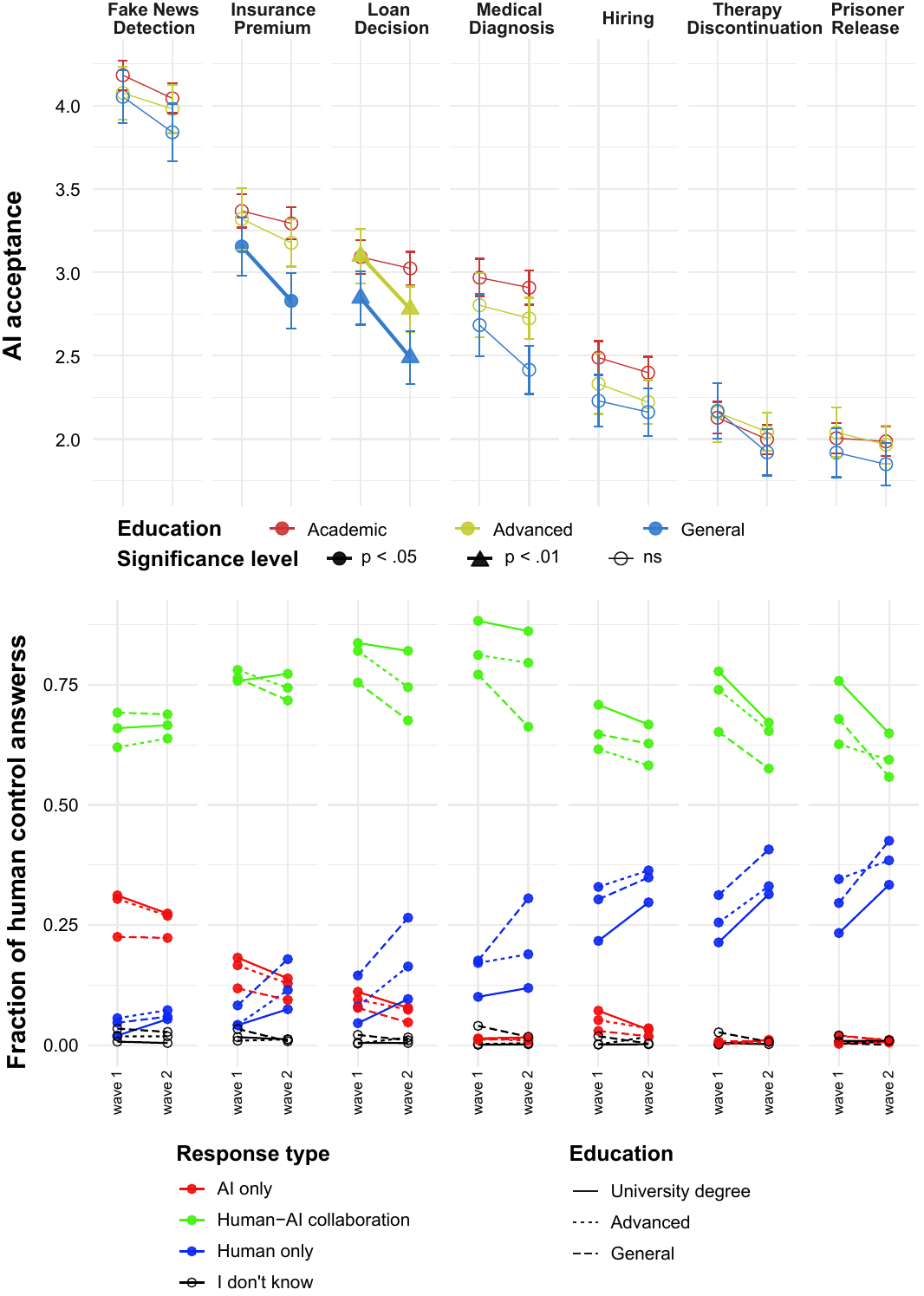}
\caption{
AI acceptance (top) and human control (bottom) across different education levels, grouped by scenarios and survey waves.
Individuals with lower education levels exhibit a larger skepticism toward AI involvement. This gap has increased with the generative AI boom: AI acceptance increased, and the perceived necessity of human involvement decreased for people with general education while remaining more stable for those with a university degree.
}
\label{fig:education_gap}
\end{figure*}

Figure~\ref{fig:education_gap} demonstrates that higher levels of education correlate with greater AI acceptance and a greater acceptance towards human-AI collaboration instead of human-only decision-making without the help of AI.

People with university degrees consistently show higher AI acceptance than those with advanced non-university education, who, in turn, exhibit more acceptance than those with only general education. This trend is evident across all surveyed scenarios. Importantly, the gap in AI acceptance between these educational groups has widened post-GenAI boom.
In particular, people with a general education level have become more critical. This is visible in the top panel of Figure~\ref{fig:education_gap} with the steeper slopes of the green lines, indicating significantly reduced acceptance levels for several scenarios such as fake news detection, insurance premiums, loan decisions, medical diagnosis, and therapy discontinuation.
For example, in the context of medical diagnosis, the acceptance for individuals with general education dropped from 2.68 to 2.40, while university-educated maintained relatively stable acceptance levels (2.97 to 2.92).

The negative regression estimates for advanced and general education in Table~\ref{tab:acceptance_glm_results_by_different_categories} show that, within both waves, a lower educational level can be predictive of lower AI acceptance.
In wave 2, general education is a significant predictor in all scenarios except therapy discontinuation).

The preference for human-AI collaboration declined across all education levels but most sharply among those with general education.
The bottom panel of Figure~\ref{fig:education_gap} illustrates that individuals with lower education levels increasingly favor human-only decision-making.
These differences were amplified with the generative AI boom, especially for loan decisions and medical diagnosis, as well as for insurance premiums.
Before the AI boom, 17.6\% of respondents with general education preferred human-only involvement in medical diagnosis, rising to 30.9\% post-boom. 
The regression coefficients in Table~\ref{tab:human_control_glm_results_by_different_categories} additionally show that lower educational level is a predictive factor for more human control after the generative AI boom, with significant positive estimates for general education across most scenarios in wave 2.

These results highlight that educational disparities have been amplified in the wake of the generative AI boom, with lower-educated individuals showing increased skepticism and a stronger preference for human oversight.

\subsubsection{Amplified language region gap.}
\label{sssec:Amplified_language_gap}

\begin{figure*}[thb]
\centering
\includegraphics[width=0.9\textwidth]{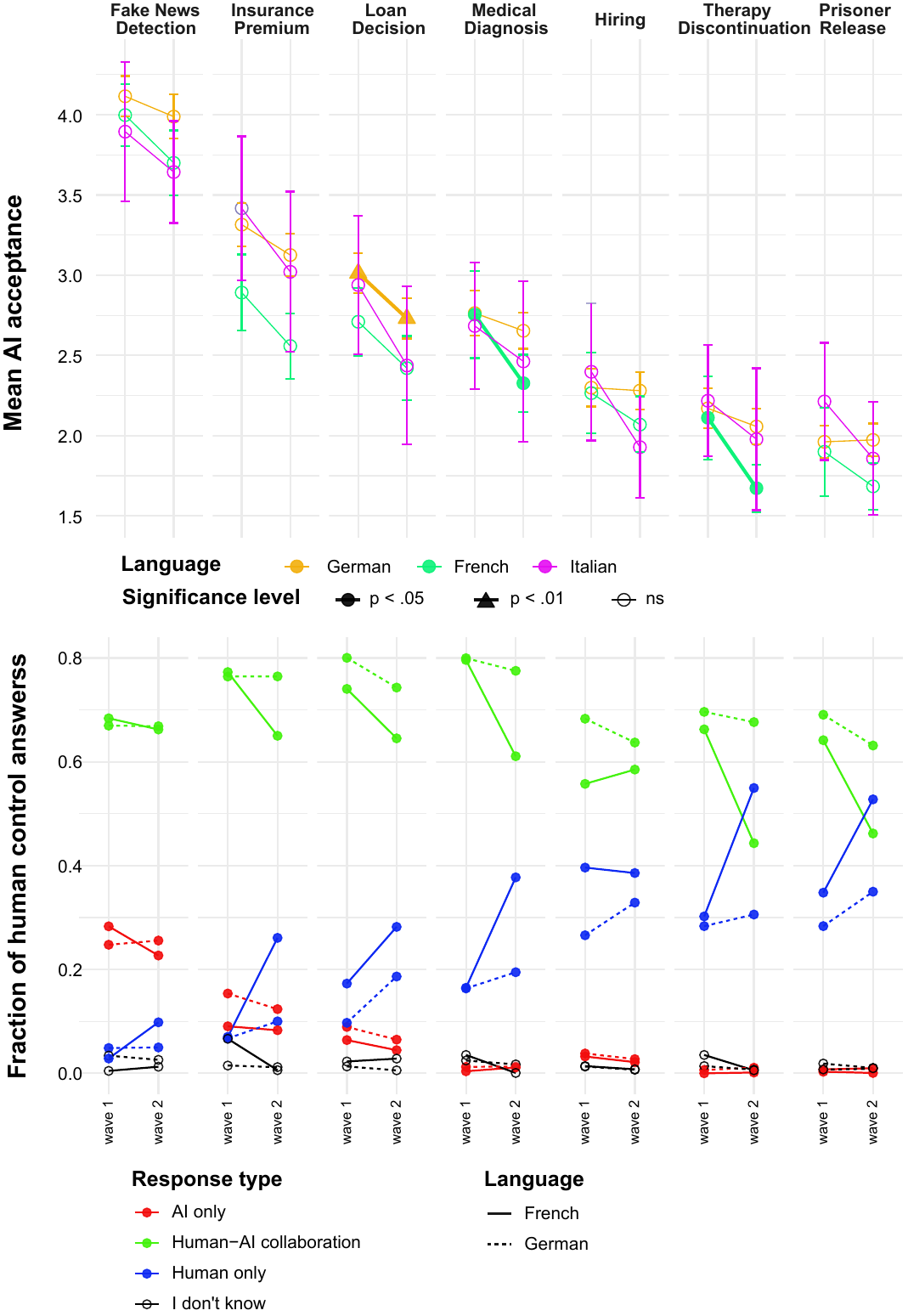}
\caption{
AI acceptance (top) and human control (bottom) across different language regions, grouped by scenarios and survey waves.
The gap between German- and French-speaking regions has increased: more skepticism toward AI among French speakers after the generative AI boom (i.e., lower AI acceptance and a shift from human-AI collaboration to human-only decisions).
}
\label{fig:language_gap}
\end{figure*}

Figure~\ref{fig:language_gap} illustrates a pronounced disparity in AI acceptance and the necessity for human control between the German-speaking and French-speaking populations in Switzerland. French speakers consistently exhibit lower AI acceptance compared to German speakers, with the gap being most pronounced in scenarios like insurance premiums, loan decisions, and the two medical scenarios.

While the disparity in AI acceptance has remained constant for most scenarios, it has widened post-GenAI boom for the two medical scenarios. For example, in the context of medical diagnosis, average AI acceptance among French speakers dropped significantly (from 2.75 to 2.33, $p<0.05$), while for German speakers, it decreased non-significantly (from 2.76 to 2.63, $p>0.05$). Similar trends are observed in therapy discontinuation, where the acceptance for French speakers fell significantly from 2.11 to 1.68 ($p<0.05$), compared to a non-significant decline from 2.17 to 2.03 among German speakers (see top panel of Figure~\ref{fig:language_gap}.
These scenarios show the same trend regarding the reported necessity of human control, as visible in the bottom panel of Figure~\ref{fig:language_gap}.

Additionally, the gap between the two language regions in terms of human control of AI-based applications also widened for the two scenarios on insurance premiums and prisoner release:
French speakers reported much less willingness of human-AI collaborations and instead started favoring more human-only settings with the generative AI boom (as visible with the dashed lines showing substantial differences between waves 1 and 2).
In contrast, German speakers only slightly increased their preference for human-only decisions post-boom.
For instance, the preference for human-only decisions in therapy discontinuation among French speakers increased from 30.17\% in wave 1 to 56.45\% in wave 2 (compared to a small increase from 28.37\% to 31.55\% for German speakers). One exception is the hiring scenario, where the preference for human control among French speakers remained stable, unlike other scenarios where significant shifts were observed.

Tables~\ref{tab:acceptance_glm_results_by_different_categories} and~\ref{tab:human_control_glm_results_by_different_categories} further highlights these trends. The regression estimates indicate significant negative coefficients for AI acceptance (positive coefficients for human control, respectively) for French speakers in wave 2 across almost all scenarios.

The trend for reported human control preferences among Italian speakers evolved similarly across both survey waves, mirroring the pattern observed among French speakers.
Therefore, for clarity, we omit it from bottom panel of Figure~\ref{fig:language_gap}.

\subsubsection{Amplified gender gap in health scenarios.}
\label{sssec:Amplified_gender_gap}

\begin{figure*}[thb]
\centering
\includegraphics[width=0.9\textwidth]{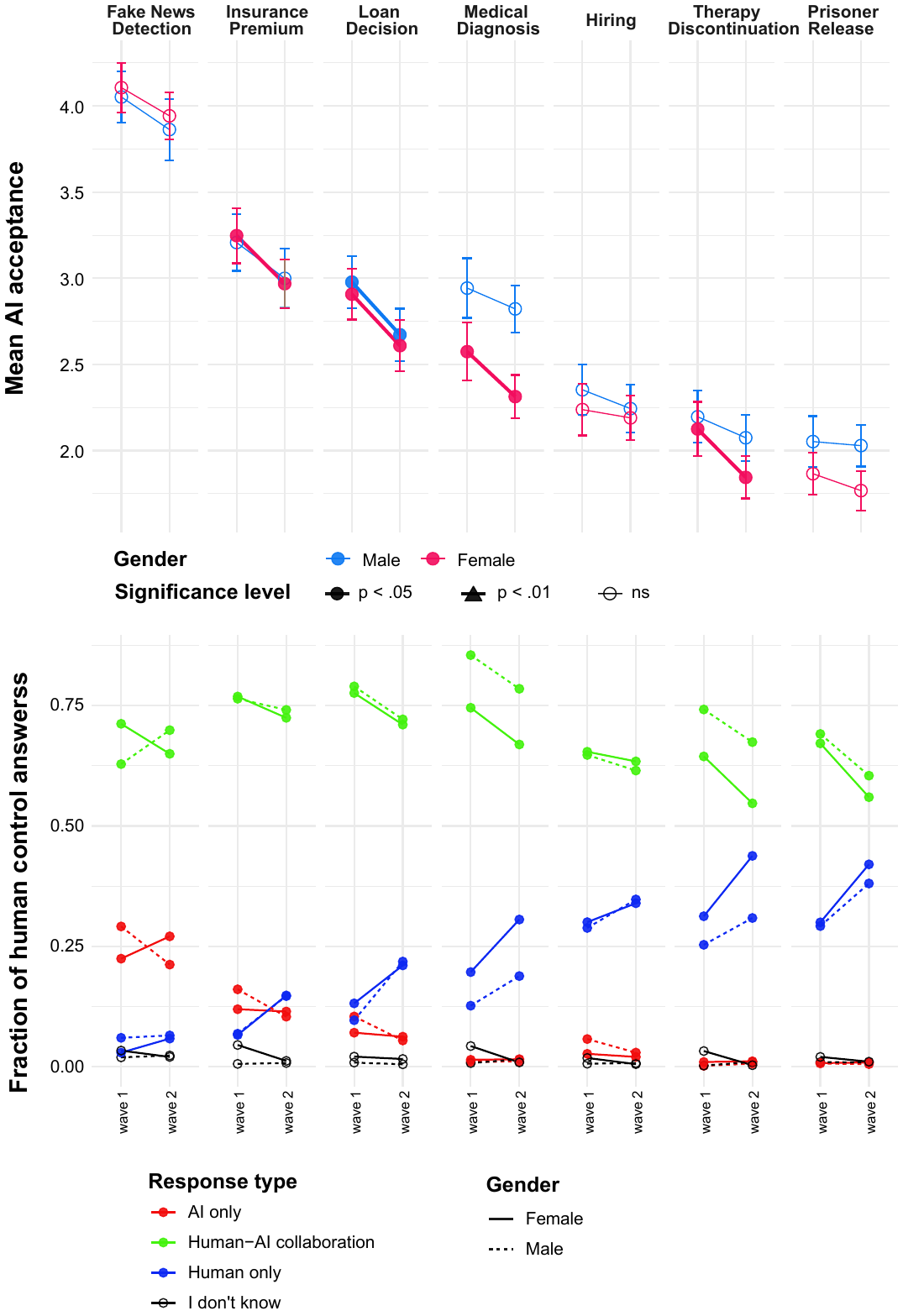}
\caption{
AI acceptance (top) and human control (bottom) across genders, grouped by scenarios and survey waves.
Women exhibit significantly lower AI acceptance compared to men, with the gap widening post-GenAI boom for health-related scenarios.
Compared to survey wave 1, women reported human-only decisions as the preferred option more often after the generative AI boom, on average.
}
\label{fig:gender_gap}
\end{figure*}

Figure~\ref{fig:gender_gap} highlights the disparities in AI acceptance and the need for human control between genders, particularly in health-related scenarios.
Women exhibit significantly more skepticism regarding the use of AI for medical diagnosis and therapy discontinuation, and this gender gap has widened post-GenAI boom.
For instance, the acceptance of AI in medical diagnosis for women decreased from 2.57 in wave 1 to 2.32 in wave 2, whereas for men, it showed a smaller decline from 2.94 to 2.80 (top panel of Figure~\ref{fig:gender_gap}). Similarly, in the context of therapy discontinuation, the acceptance among women dropped from 2.13 to 1.83, while for men, it decreased slightly from 2.20 to 2.07.
In the therapy discontinuation scenario, men and women had similarly low acceptance on average in wave 1 (2.20 for men and 2.13 for women).
With the generative AI boom, this decreased slightly for men (to 2.07), but women became significantly less accepting of using AI for therapy discontinuation decisions (to 1.83).
This trend is further supported by the regression estimates in Table~\ref{tab:acceptance_glm_results_by_different_categories}, which show significant negative coefficients for women in scenarios like medical diagnosis and therapy discontinuation in wave 2.
Furthermore, women also showed a lower acceptance for using AI in prisoner release scenarios in both waves -- see significant negative regression estimate for Gender (f) in Table~\ref{tab:acceptance_glm_results_by_different_categories}.

The bottom panel of Figure~\ref{fig:gender_gap} tells a similar story, showing larger increases in human-only answers among women compared to men, for the aforementioned scenarios of medical diagnosis, therapy discontinuation, and prisoner release.
Table~\ref{tab:human_control_glm_results_by_different_categories} supports these observations, showing significant positive coefficients for women in wave 2 for the scenarios of medical diagnosis and therapy discontinuation.

\begingroup
\setlength{\tabcolsep}{3pt}
\begin{table*}[thb]
\caption{Predictors of AI acceptance in different scenarios.} 
\label{tab:acceptance_glm_results_by_different_categories}
\scriptsize
\centering
\begin{tabular}{lccccccc} 
\multicolumn{5}{l}{{\normalsize Dependent variable: \textbf{AI acceptance}}} \vspace{1mm}\\
  \toprule
 \textit{{\small Scenario $\;\;\rightarrow$}} & \makecell{\textbf{Medical} \\ \textbf{Diagnosis}} & \makecell{\textbf{Therapy} \\ \textbf{Discontinuation}} & \makecell{\textbf{Fake News} \\ \textbf{Detection}} & \makecell{\textbf{Hiring}} & \makecell{\textbf{Loan} \\ \textbf{Decision}} & \makecell{\textbf{Insurance} \\ \textbf{Premium}} & \makecell{\textbf{Prisoner} \\ \textbf{Release}} \\ %
  \midrule
  \multicolumn{5}{l}{\textbf{{\small$\!\!\!$By education level}}}  \\
  \textbf{\textit{Wave 1}} &&&&&&& \\\rowcolor{LightGray}
  Advanced education & -0.17 (0.11) & 0.03 (0.1) & -0.11 (0.09) & -0.16 (0.11) & 0.01 (0.1) & -0.05 (0.11) & 0.03 (0.09)\\ \rowcolor{LightGray}%
  General education & \textbf{-0.28 (0.11)*} & 0.04 (0.1) & -0.13 (0.09) & \textbf{-0.26 (0.09)*} & \textbf{-0.25 (0.1)*} & -0.21 (0.1) & -0.09 (0.09)\\  %
  \textbf{\textit{Wave 2}} &&&&&&& \\\rowcolor{LightGray}
  Advanced education & -0.18 (0.08) & 0.04 (0.07) & -0.06 (0.09) & -0.18 (0.08) & \textbf{-0.24 (0.09)*} & -0.12 (0.09) & -0.02 (0.07) \\ \rowcolor{LightGray} %
  General education & \textbf{-0.49 (0.09)***} & -0.08 (0.08) & -0.2 (0.1) & \textbf{-0.24 (0.09)*} & \textbf{-0.53 (0.1)***} & \textbf{-0.46 (0.1)***} & -0.14 (0.08) \\ %
  \midrule
  \multicolumn{5}{l}{\textbf{{\small$\!\!\!$By language region}}}  \\
  \textbf{\textit{Wave 1}} &&&&&&& \\\rowcolor{LightGray}
  French language & -0.01 (0.16) & -0.06 (0.15) & -0.12 (0.12) & -0.03 (0.14) & \textbf{-0.3 (0.13)*} & \textbf{-0.42 (0.14)**} & -0.06 (0.15) \\ \rowcolor{LightGray}%
  Italian language & -0.08 (0.21) & 0.05 (0.19) & -0.22 (0.23) & 0.1 (0.23) & -0.07 (0.23) & 0.1 (0.24) & 0.25 (0.19) \\   %
  \textbf{\textit{Wave 2}} &&&&&&& \\  \rowcolor{LightGray}
  French language & \textbf{-0.33 (0.11)**} & \textbf{-0.38 (0.09)***} & \textbf{-0.29 (0.12)*} & -0.21 (0.11) & \textbf{-0.31 (0.12)*} & \textbf{-0.57 (0.12)***} & \textbf{-0.29 (0.09)**} \\ \rowcolor{LightGray}  %
  Italian language & -0.19 (0.26) & -0.08 (0.23) & -0.35 (0.18) & -0.35 (0.17) & -0.29 (0.26) & -0.1 (0.26) & -0.12 (0.19) \\ %
  \midrule
  \multicolumn{5}{l}{\textbf{{\small$\!\!\!$By gender}}}  \\
  \textbf{\textit{Wave 1}} &&&&&&& \\\rowcolor{LightGray}
  Gender (f) & \textbf{-0.37 (0.12)**} & -0.07 (0.11) & 0.05 (0.11) & -0.12 (0.11) & -0.07 (0.11) & 0.04 (0.12) & -0.19 (0.1)\\ %
  \textbf{\textit{Wave 2}} &&&&&&& \\ \rowcolor{LightGray}
  Gender (f) & \textbf{-0.51 (0.09)***} & \textbf{-0.23 (0.09)*} & 0.08 (0.11) & -0.05 (0.1) & -0.06 (0.11) & -0.03 (0.11) & \textbf{-0.26 (0.09)**}\\ 
   \bottomrule
\end{tabular}
\begin{tablenotes}
  \small
  \item Note: This table summarizes the results of 42 separate survey-weighted linear regression models, which cover 7 scenarios, 3 demographic attributes, and 2 survey waves. The dependent variable indicates the acceptance of using AI (1=not acceptable at all, ..., 5=definitely acceptable).
  Each cell presents regression estimates with standard errors in parentheses. Significant predictors are indicated in bold, with significance levels as follows: *** $p<0.001$, ** $p<0.01$, * $p<0.05$.
\end{tablenotes}
\end{table*}
\endgroup

\begingroup
\setlength{\tabcolsep}{3.2pt}
\begin{table*}[thb]
\caption{Predictors of the need for human control in different scenarios.} 
\label{tab:human_control_glm_results_by_different_categories}
\scriptsize
\centering
\begin{tabular}{lccccccc} %
\multicolumn{5}{l}{{\normalsize Dependent variable: \textbf{human control}}} \vspace{1mm}\\
  \toprule
 \textit{{\small Scenario $\;\;\rightarrow$}} & \makecell{\textbf{Medical} \\ \textbf{Diagnosis}} & \makecell{\textbf{Therapy} \\ \textbf{Discontinuation}} & \makecell{\textbf{Fake News} \\ \textbf{Detection}} & \makecell{\textbf{Hiring}} & \makecell{\textbf{Loan} \\ \textbf{Decision}} & \makecell{\textbf{Insurance} \\ \textbf{Premium}} & \makecell{\textbf{Prisoner} \\ \textbf{Release}} \\ %
  \midrule
  \multicolumn{5}{l}{\textbf{{\small$\!\!\!$By education level}}}  \\
  \textbf{\textit{Wave 1}}  \\\rowcolor{LightGray}
  Advanced education & 0.05 (0.06) & 0.06 (0.05) & 0.14 (0.07) & \textbf{0.19 (0.07)*} & 0.08 (0.06) & 0.1 (0.07) & 0.06 (0.06) \\ \rowcolor{LightGray}%
  General education & 0.11 (0.05) & 0.1 (0.06) & \textbf{0.23 (0.06)***} & \textbf{0.21 (0.06)**} & \textbf{0.22 (0.07)**} & \textbf{0.21 (0.06)**} & 0.04 (0.06) \\ %
  \textbf{\textit{Wave 2}}  \\\rowcolor{LightGray}
  Advanced education & 0.09 (0.05) & 0.01 (0.05) & 0.07 (0.06) & 0.11 (0.06) & \textbf{0.17 (0.05)**} & 0.11 (0.06) & 0.06 (0.05) \\ \rowcolor{LightGray}%
  General education & \textbf{0.23 (0.06)***} & 0.07 (0.05) & 0.12 (0.06) & 0.09 (0.06) & \textbf{0.36 (0.06)***} & \textbf{0.32 (0.06)***} & 0.1 (0.05) \\ 
  \midrule
  \multicolumn{5}{l}{\textbf{{\small$\!\!\!$By language region}}}  \\
  \textbf{\textit{Wave 1}}  \\\rowcolor{LightGray}
  French language & 0 (0.08) & -0.01 (0.08) & 0.04 (0.1) & 0.14 (0.09) & 0.21 (0.11) & 0.18 (0.09) & 0.06 (0.09) \\ \rowcolor{LightGray}%
  Italian language & -0.1 (0.12) & -0.14 (0.1) & 0.01 (0.15) & -0.2 (0.19) & -0.02 (0.16) & 0.05 (0.14) & -0.17 (0.14) \\
  \textbf{\textit{Wave 2}}  \\\rowcolor{LightGray}
  French language & \textbf{0.3 (0.07)***} & \textbf{0.33 (0.06)***} & 0.11 (0.08) & 0.13 (0.07) & \textbf{0.2 (0.08)*} & \textbf{0.37 (0.09)***} & \textbf{0.23 (0.06)***} \\\rowcolor{LightGray}
  Italian language & 0.25 (0.15) & 0.21 (0.16) & \textbf{0.37 (0.12)**} & \textbf{0.24 (0.09)*} & \textbf{0.34 (0.13)*} & \textbf{0.5 (0.15)**} & 0.1 (0.17) \\
  \midrule
  \multicolumn{5}{l}{\textbf{{\small$\!\!\!$By gender}}}  \\
  \textbf{\textit{Wave 1}}  \\\rowcolor{LightGray}
  Gender (f) & 0.12 (0.06) & 0.1 (0.06) & 0.08 (0.07) & 0.07 (0.07) & 0.09 (0.08) & 0 (0.07) & -0.01 (0.06) \\ %
  \textbf{\textit{Wave 2}}  \\ \rowcolor{LightGray}
  Gender (f) & \textbf{0.18 (0.06)*}* & 0.13 (0.06) & -0.12 (0.07) & 0.02 (0.06) & -0.09 (0.07) & 0 (0.07) & 0.02 (0.06) \\ %
   \bottomrule
\end{tabular}
\begin{tablenotes}
  \small
  \item Note: This table summarizes the results of 42 separate survey-weighted linear regression models, encompassing 7 scenarios, 3 demographic attributes, and 2 survey waves. The dependent variable indicates the preference for decision-making between humans, AI, or both together (1=human only, ..., 4=AI only).
  Each cell presents regression estimates with standard errors in parentheses. Significance levels are denoted as follows: *** $p<0.001$, ** $p<0.01$, * $p<0.05$.
\end{tablenotes}
\end{table*}
\endgroup

\end{document}